\documentclass{elsarticle}

\usepackage{framed,multirow}

\usepackage{amssymb}
\usepackage{latexsym}

\usepackage{bm}
\usepackage{amsmath}
\usepackage{graphicx}

\usepackage{adjustbox}
\usepackage{booktabs}
\usepackage{multirow}
\usepackage[table]{xcolor}
\usepackage{lscape}
\usepackage{array}
\usepackage{graphicx}
\usepackage{geometry}

\usepackage{url}
\usepackage{bm}
\usepackage{xcolor}
\usepackage{float}
\usepackage{amsmath}
\usepackage{array}
\usepackage{pgfplots} 

\newcommand{\R}{\mathbb{R}} 
\newcommand{\E}{\mathbb{E}} 
\newcommand{\N}{\mathcal{N}} 
\newcommand{\G}{\mathbf{\Gamma}}

\DeclareMathOperator{\X}{\mathbf{X}}

\DeclareMathOperator{\xn}{\mathbf{x}_{n,:}}

\DeclareMathOperator{\xnd}{x_{n,d}}
\DeclareMathOperator{\xnt}{\mathbf{x}_{n,:}^\top}

\DeclareMathOperator{\xlam}{\Lambda_{\mathbf{x_{n,:}}}}

\DeclareMathOperator{\Xs}{\mathbf{\tilde{X}}}
\DeclareMathOperator{\Xst}{\mathbf{\tilde{X}^\top}}

\DeclareMathOperator{\xsd}{\mathbf{\tilde{x}}_{:,d}}
\DeclareMathOperator{\xsdt}{\mathbf{\tilde{x}}_{:,d}^\top}

\DeclareMathOperator{\as}{\mathbf{a}}

\DeclareMathOperator{\an}{a_{\hat{n}}}

\DeclareMathOperator{\aE}{\langle \mathbf{a}\rangle}
\DeclareMathOperator{\aat}{\langle \mathbf{a}\mathbf{a}^\top\rangle}

\DeclareMathOperator{\vs}{\mathbf{v}}

\DeclareMathOperator{\vd}{v_{d}}
\DeclareMathOperator{\vlam}{\Lambda_{\mathbf{v}}}
\DeclareMathOperator{\vE}{\langle \mathbf{v}\rangle}
\DeclareMathOperator{\vdE}{\langle v_{d}\rangle}
\DeclareMathOperator{\vvt}{\langle \mathbf{v}\mathbf{v}^\top\rangle}
\DeclareMathOperator{\vlamE}{\Lambda_{\langle \mathbf{v}\rangle}}

\DeclareMathOperator{\bb}{b}

\DeclareMathOperator{\ys}{\mathbf{y}}
\DeclareMathOperator{\yn}{y_{n}}

\DeclareMathOperator{\A}{\mathbf{a}}

\DeclareMathOperator{\apsin}{\alpha_{\tilde{n}}^{\psi}}
\DeclareMathOperator{\bpsin}{\beta_{\tilde{n}}^{\psi}}

\DeclareMathOperator{\fin}{\psi_{n}}
\DeclareMathOperator{\fis}{\bm{\psi}}

\DeclareMathOperator{\alfad}{\delta_{d}}
\DeclareMathOperator{\alfas}{\bm{\delta}}

\DeclareMathOperator{\acov}{\Sigma_{\mathbf{a}}}

\DeclareMathOperator{\anfi}{\alpha_{n}^{\psi}}
\DeclareMathOperator{\bnfi}{\beta_{n}^{\psi}}
\DeclareMathOperator{\filam}{\Lambda_{\bm{\psi}}}
\DeclareMathOperator{\filamE}{\Lambda_{\langle \bm{\psi}\rangle}}

\DeclareMathOperator{\adalfa}{\alpha_{d}^{\delta}}
\DeclareMathOperator{\bdalfa}{\beta_{d}^{\delta}}
\DeclareMathOperator{\aceroalfa}{\alpha_{0}^{\delta}}
\DeclareMathOperator{\bceroalfa}{\beta_{0}^{\delta}}
\DeclareMathOperator{\aalfa}{\bm{\alpha}^{\delta}}
\DeclareMathOperator{\balfa}{\bm{\beta}^{\delta}}

\DeclareMathOperator{\acerofi}{\alpha_{0}^{\psi}}
\DeclareMathOperator{\bcerofi}{\beta_{0}^{\psi}}
\DeclareMathOperator{\afi}{\bm{\alpha}^{\psi}}
\DeclareMathOperator{\bfi}{\bm{\beta}^{\psi}}

\DeclareMathOperator{\atau}{\alpha^{\tau}}
\DeclareMathOperator{\btau}{\beta^{\tau}}
\DeclareMathOperator{\acerotau}{\alpha_{0}^{\tau}}
\DeclareMathOperator{\bcerotau}{\beta_{0}^{\tau}}
\DeclareMathOperator{\taus}{\tau}
\DeclareMathOperator{\tauE}{\langle\tau\rangle}

\DeclareMathOperator{\unmed}{\frac{1}{2}}
\DeclareMathOperator{\lnpi}{\ln(2\pi)}
\DeclareMathOperator{\lntau}{\ln(\tau)}

\DeclareMathOperator{\sumn}{\sum_{n=1}^{N}}
\DeclareMathOperator{\sumns}{\sum_{\hat{n}}^{\hat{N}}}
\DeclareMathOperator{\sumd}{\sum_{d=1}^{D}}
\DeclareMathOperator{\sumduno}{\sum_{d1=1}^{D}}
\DeclareMathOperator{\sumddos}{\sum_{d2=1}^{D}}
\DeclareMathOperator{\sumdrara}{\sum_{\hat{d}=1}^{\hat{D}}}

\DeclareMathOperator{\bE}{\langle b\rangle}
\DeclareMathOperator{\sigd}{\sigma_{d}}

\DeclareMathOperator{\y}{\mathbf{y}}
\DeclareMathOperator{\yas}{y_{*}}

\DeclareMathOperator{\Xn}{\mathbf{x}_{n,:}} 

\DeclareMathOperator{\Aa}{\mathbf{a}} 

\usepackage{hyperref}

\usepackage[normalem]{ulem} 

\definecolor{bblue}{HTML}{4F81BD}
\definecolor{rred}{HTML}{C0504D}
\definecolor{ggreen}{HTML}{9BBB59}
\definecolor{ppurple}{HTML}{9F4C7C}
\definecolor{oorange}{HTML}{BD814F}
\definecolor{rrare}{HTML}{81BD4F}
\definecolor{ggray}{HTML}{BD4F81}

\usepackage{pgfplotstable} 
\usepackage{booktabs}
\usepackage{adjustbox}
\usepackage{multirow}
\usepackage{tikz}

\pgfplotsset{
/pgfplots/my legend/.style={
legend image code/.code={
\draw[thick,black](-0.05cm,0cm) -- (0.3cm,0cm);%
   }
  }
}
\definecolor{newcolor}{rgb}{.8,.349,.1}

\journal{arXiv}

\begin{document}

\begin{frontmatter}

\title{The Relevance Feature and Vector Machine for health applications}%

\author[1]{Albert Belenguer-Llorens\corref{cor1}}
\cortext[cor1]{Corresponding author. 
  }
\ead{abelengu@pa.uc3m.es}

\author[1,2]{Carlos Sevilla-Salcedo}
\author[1]{Emilio Parrado-Hernández}
\author[1]{Vanessa G\'{o}mez-Verdejo}

\address[1]{Department of Signal Processing and Communications, Universidad Carlos III de Madrid, Leganés, 28911, Spain}
\address[2]{Department of Computer Science, Aalto University, Espoo, 02150, Finland}

\begin{abstract}
This paper presents the Relevance Feature and Vector Machine (RFVM), a novel model that addresses the challenges of the fat-data problem when dealing with clinical prospective studies. The fat-data problem refers to the limitations of Machine Learning (ML) algorithms when working with databases in which the number of features is much larger than the number of samples (a common scenario in certain medical fields). To overcome such limitations, the RFVM incorporates different characteristics: (1) A Bayesian formulation which enables the model to infer its parameters without overfitting thanks to the Bayesian model averaging. (2) A joint optimisation that overcomes the limitations arising from the fat-data characteristic by simultaneously including the variables that define the primal space (features) and those that define the dual space (observations). (3) An integrated prunning that  removes the irrelevant features and samples during the training iterative optimization. Also, this last point turns out crucial when performing medical prospective studies, enabling researchers to exclude unnecessary medical tests, reducing costs and inconvenience for patients, and identifying the critical patients/subjects that characterize the disorder and, subsequently, optimize the patient recruitment process that leads to a balanced cohort. The model capabilities are tested against state-of-the-art models in several medical datasets with fat-data problems. These experimental works show that RFVM is capable of achieving competitive classification accuracies while providing the most compact subset of data (in both terms of features and samples). Moreover, the selected features (medical tests) seem to be aligned with the existing medical literature.

\end{abstract}


\end{frontmatter}

\section{Introduction}

The fat-data or low sample-to-feature ratio characteristic graphically describes the  challenges that arise when dealing with databases in  which the number of features ($D$) greatly exceeds the number of samples ($N$). This characteristic describes datasets in a broad number of domains, although this paper focuses in medical data analysis with machine learning (ML). An immediate source of fat-data in clinical  analytics is caused by the use of some techinques for medical diagnosis, such as  Magnetic Resonance Imaging (MRI), Positron Emission Tomography (PET) or genetic analysis, that generate observations with a humongous number of features. This effect is accentuated by the fact that enrolling patients and control subjects in clinical studies remains a challenge due to legal and administrative protocols or the rarity of certain diseases.  The usual approach to build ML models with the outcome of these data acquisition processes is to first preprocess the data with feature selection (FS) \cite{bolon2015feature,garcia2023evolutionary} or feature extraction \cite{zebari2020comprehensive} techniques in order to reduce the dimensionality of the data to levels that are appropriate for the ML model that constitutes the core of the  analysis. Another very interesting framework for the application of the ideas developed in this paper is the case of prospective studies, in which researchers collect data over extended periods of time alongside ongoing clinical studies. The management of these prospective studies would certainly benefit from an adaptive refinement of the data acquisition process. In this context adaptive refinement means that the researchers could start with a  more or less exhaustive set of medical tests, and to adaptively reduce its number as the data analytics unveil the relevance, redundancy or irrelevance of some of those tests for the characterization of the  disesase under study. Moreover, the cohort of subjects could also be refined, removing subjects if the researchers detect redundancies (in terms of information relevant for the model), or recruiting new subjects when some regions of the population are  perceived as underrepresented. Controlling the number of features (specially medical tests, but also socio/demographic features, clinical history, etc.) helps control the sustainability of the study, and what is more important, the patient discomfort. As stated before, FS techniques can help identify this compact set of medical tests needed to feed the models. For the identification of relevant subjects, one can resource to the concept of sparsity in ML models. Sparsity means that the architecture of the model is defined in terms of a subset of input observations, such as the support vectors (SVs) in the Suport Vector Machine (SVM) \cite{cristianini2007support}, the relevance vectors (RVs) in the Relevance Vector Machine (RVM) \cite{tipping1999relevance}, the inducing points in the sparse Gaussian Processes (SGPs) \cite{rasmussen2003gaussian},  or the centroids in the KMeans \cite{sinaga2020unsupervised}.  The fact that these methods rely on a subset of critical samples to determine the model architecture can help assess the importance of each sample in the training data set as a container of information relevant for the characterization of the task at hand, information that should be captured by the model. Therefore, achieving two-way sparsity (sparsity in the domain of the features and in the domain of the subjects) becomes a potentially very beneficial design objective for the ML models that would eventually work in those scenarios.

An immediate way of achieving dual sparsity within supervised learning is the sequential combination of a  FS technique followed by the fitting of a sparsity-enforcing method, such as the aforementioned SVM, SGP or RVM. This approach usually ends up delivering suboptimal models, as both optimizations (the selection of the relevant features and the fitting of the sparse model with the compact set of features) are addressed independently. Other state-of-the art techniques look to obtain this dual sparsity by merging both goals in a same optimization (more details in Section \ref{sec:Related}). However, those models described in the related work are not particularly tailored to face fat-data situations, and also fail to deliver acceptable solutions due to diverse reasons, such as having to deal with large covariance matrices, or presenting problems with the convergence of their loss functions. In this sense, the proposal of this paper aims at overcoming those limitations. The model here presented is named \textbf{Relevance Feature and Vector Machine} (RFVM), as it aims at fitting an accurate ML model while achieving  two-way sparsity.  With respect to the sparsity in the data samples, RFVM follows the RVM philosophy rather than the SVM one. Despite the RVM is the Bayesian formulation of the SVM, it selects RVs instead of SVs to build the model architecture. In a classification setting, the  SVs are observations whose main purpose is the definition of classification boundaries. These SVs lie close to the boundaries and, in cases in which the output classes highly overlap, all the training observations that are misclassified become SVs. On the contrary,  RVs extend the concept of relevance to the whole input space, not just to the training observations lying close to the boundary. Hence, as the objective underlying sparsity in the dual space is to assess the relevance of the observations in order to guide the management of the subjects cohort in a clinical study, we stick to the more general RV concept.

RFVM combines several advantageous capabilities to excel in the delivery of ML models in fat-data scenarios, like those faced in clinical data analytics. First, it employs a Bayesian formulation that provides with robust and comprehensive solutions in problems with a reduced set of observations\cite{winschel2010solving,chandra2023escaping,koop2016bayesian}.  Second, RFVM simultaneously infers the relevance of the features and of the observations within a single iterative process that exploits the connections between them. Finally, it is remarkable that the feature relevance inference is formulated in the dual space (with $N$ variables, one per observation), what represent significant computational advantages in fat-data problems.

The remainder of the paper is organized as follows. Section \ref{sec:Related} reviews several classical and state-of-the-art ML approaches to address the challenges associated with the fat-data problem. Section \ref{sec:Model} introduces the formulation of the RFVM, presenting the generative model, the inference process and the predictive distribution. Section \ref{sec:Results} assesses the performance of  RFVM, considering factors such as accuracy, FS, and Relevance Vector Selection (RVS), in comparison to multiple baseline models. This section also includes a clinical analysis of the features obtained by RFVM in a well-studied and interpretable dataset, and a computational cost analysis. Finally, Section \ref{sec:Conclusions} summarizes our findings and draws the main conclusions of this work.

\section{Related work}
\label{sec:Related}

In the realm of ML, several models have been proposed to address the  two-way sparsity challenge. On the one hand, the classical approach is to conduct a previous FS before fitting a sparse-inducing final model. The range of FS methods that somehow align the FS with the final model fitting (although through separate optimizations) are traditionally clustered into two main families: wrapper methods and  embedded methods.  Recursive Feature Elimination \cite{chen2007enhanced} (RFE) is perhaps one of the most used techinques within the former.  RFE is a greedy iterative procedure that at each iteration removes the feature found to be the least relevant amongst all the survivors up to that iteration. RFE can be applied to any ML model as the model is used as a black-box, there is no need to introduce any modification to the model original formulation. Embedded methods englobe those cases in which the optimization that happens during the fitting of some ML models provides with levers that enforce feature selection, for instance with regularizations that induce sparsity, such as the L1 penalty of LASSO  regression \cite{tibshirani1996regression}, the Logistic regression with elastic-net \cite{algamal2015high} or the group LASSO \cite{simon2012standardization}, which extends the FS capabilities to the multioutput scenario. After the FS has taken place, the data set with the reduced dimensionality is processed with a sparsity-inducing ML model, that takes care of the selection of the relevant input observations. Examples of such sparsity-inducing models are the previously mentioned SVMs, SGPs or RVMs.

The suboptimality arising from the decoupling of the FS and the fitting of the main model can be overcome by resourcing to methods that address both sparsities within a same optimization. The SVM-$\ell$1 \cite{zhu20031}, for instance,  stands out for its ability to merge support vector selection (SVS) characteristics with FS stemming from L1 regularization. However, this combination introduces certain limitations, including challenges in optimization, excessive feature sparsity in specific scenarios, and low data sparsity \cite{dedieu2022solving, zou2005regularization}.

 In the Bayesian ML paradigm there are also examples of models that separately address the sparsity either in the features or in the data samples  \cite{murphy2012machine}. These techniques directly introduce priors into the model variables  that nullify the least important components. Although there are several specific priors for this purpose, the most commonly adopted  one is the Automatic Relevance Determination (ARD) prior. In the ARD paradigm one assigns a gaussian  prior with 0 mean and precision equal to $\eta_i$  to each variable $\theta_i$ in the model. Moreover, each $\eta_i$ itself is assigned a prior gamma distribution with its mode located at 0. The application of ARD to the features serves as the foundation for the FS of the Bayesian linear regression \citep{bishop2006pattern}, while its application to the dual variables results in the Sparse Relevance Vector Machine (SRVM) \cite{tipping2001sparse}, which is the Bayesian formulation of the SVM.

Furthermore, within the Bayesian models, there are two approaches that seem to stand out in addressing the challenge of combining sparsity over features and samples within the same optimization. First, we highlight the combination of the SGPs \citep{rasmussen2003gaussian} with an ARD prior. In this case, the formulation of the SGP sets sparsity over the data points by expressing the posterior distribution over a subset of  input observations known as inducing points\footnote{Inducing points are used within the SGP formulation as a reduced set of representative points that captures the original data structure in order to reduce the complexity and computational cost of the training and test procedure.}. At the same time, the ARD prior  sets the relevances/weights of each variable in the model, emphasizing the significance of each feature rather than actively selecting them. The second approach is the Probabilistic Feature Selection and Classification Vector Machine (PFCVM) \cite{jiang2019probabilistic}. This model manages both feature and RV selection within the same iterative procedure. However, it may not be ideally suited for addressing scenarios involving datasets with an excessive number of features, as it requires handling large covariance matrices, introducing severe computational and parameter inference challenges in the large-dimensional setting.

\section{The proposed model: The RFVM}
\label{sec:Model}

This section introduces the Relevance Feature and Vector Machine (RFVM), a novel Bayesian ML model designed to efficiently yield sparsity in terms of both features and RVs for fat-data problems.

Let us consider a data set with $N$ observations and their corresponding targets, $\{(\Xn, t_n)\}_{n=1}^N$, where $\Xn \in \R^D$ and $t_n \in \{0,1\}$. Furthermore, let us denote by $\X$ the $N\times D$ matrix with the stacked training data, where $\Xn$ and $\mathbf{x}_{:,d}$  denote its rows and columns, respectively, and $\mathbf{t}$ the $N$-dimensional vector that contains the $t_n$ targets; furthermore, $diag(\mathbf{X})$ and $Trace(\mathbf{X})$ denote the diagonal and the trace of $\mathbf{X}$, respectively. Also, for the r.v., $\langle \mathbf{z}\rangle$ and $\Sigma_{\mathbf{z}}$ denote the mean and covariance of $\mathbf{z}$, respectively. Identity matrices of size $N$ are denoted by $\mathbf{I}_{N}$, and $\Lambda_{\mathbf{\mathbf z}}$ is the diagonal matrix formed  by vector $\mathbf{z} = [z_1, ..., z_P]$ so that $[\Lambda_{\mathbf{\mathbf z}}]_{p,p} = z_p$, also, $\mathbf{1}_{N}$ denotes a $N$ dimensional vector filled by 1's. Furthermore,    $\Xs \in \R^{\tilde{N} \times D}$ is the matrix that contains the $\tilde{N}$ data observations candidate to be RVs. Notice that this set of candidates usually comprises the whole training dataset; however, it could also contain a subset of it, a new set of data points different from the training data, or a combination of training observations and new data points. Although in the experiments $\Xs = \X$, this  notation will be kept throughout the mathematical developments to define generalizable equations, and also to be able to distinguish which specific parts of our formulation affect to either the input data or the RVs. Finally, we denote with $\chi_{\mathbf{z}}$ the domain of the r.v. $\mathbf{z}$ when integrating over a certain function $F(\mathbf{z})$, i.e., $\int_{\chi_{\mathbf{z}}}F(\mathbf{z})d\mathbf{z}$.

\subsection{Generative Model}
\label{sec:generative}

First of all, within the RFVM, we follow the standard Bayesian procedure when defining a binary classification framework. That is, we defined a latent continuous variable $\mathbf{y}$ which is related with the class label $\mathbf{t}$ through a Bayesian logistic regression
\begin{equation}
    p(t_n|\yn) =  \sigma(\yn)^{t_n}(1-\sigma(\yn))^{1-t_n} =   e^{\yn t_n}\sigma(-\yn), \quad n=1,\dots,N
    \label{eq:t}
\end{equation}
where $\sigma(x) = (1+e^{-x})^{-1}$ is the sigmoid function. Now, we can easily  cast the model as a regression problem over $\mathbf{y}$ to subsequently treat it as a binary classification by means of Eq. (\ref{eq:t}).

Hence, the RFVM model builds on the SRVM formulation, where the relationship between an input sample, $\xn$, and its corresponding latent target $\yn$ is defined as
\begin{equation}
    \yn = \xn\Xst\as + b + \epsilon,
    \label{eq:mod1}
\end{equation}
where $\mathbf a = [a_1, \ldots, a_{\tilde{N}}]$ are random variables that serve as the dual coefficients for the model. Vector $\mathbf a$ follows a multivariate gaussian distribution with $\mathbf{a} \sim \N(\mathbf 0, \Lambda_{\boldsymbol{\psi}}^{-1})$ where $\boldsymbol{\psi} = [\psi_1, \ldots, \psi_{\tilde{N}}]$ defines an ARD prior by setting a gamma distribution over each $\psi_{\tilde{n}}$ as $\psi_{\tilde{n}} \sim \G(\apsin, \bpsin)$ and subsequently imposing sparsity over the RVs.  In addition, $b \sim \N(0,1)$ corresponds to the bias term and $\epsilon$ is gaussian noise with zero-mean and precision $\tau \sim \G(\atau, \btau)$.

To endow the model with FS capabilities, we introduce an additional random vector $\mathbf v = [v_1,\ldots,v_{D}]$ in the model definition 
\begin{equation}
    \yn = \xn\vlam\Xst\as + b + \epsilon,
    \label{eq:mod2}
\end{equation}
where each element $\vd$ aims at determining which  columns of $\X$ are irrelavant. For this purpose, an ARD prior  is assigned to  each $\vd$
\begin{equation}
    \vd \sim \N_F(0,\delta_d^{-1}),
    \label{eq:nfv}
\end{equation}
where $\N_F$ is a  folded normal distribution \citep{tsagris2014folded} (more details in \ref{sec_ApFolded}) with parameter $\alfad \sim \G(\adalfa,\bdalfa)$, $d=1,\dots,D$. 
The role of the ARD priors is to force that those elements of $\mathbf v$ associated to irrelevant features become 0 during the model inference. Since the purpose of the $\{\vd\}_{d=1}^D$ is to capture the relevance of the features, they need to be positive. In other words, the distribution that models each $\vd$ must have a non-negative support. In this sense, $\N_F$ is selected over other options with non-negative support, such as the exponential or the gamma distributions, because it enables reaching analytical solutions through the mean-field variational inference procedure explained in Subsection \ref{sec:inference}.

\begin{figure}[!h]
  \begin{minipage}{.5\textwidth}
        \includegraphics[width=\linewidth]{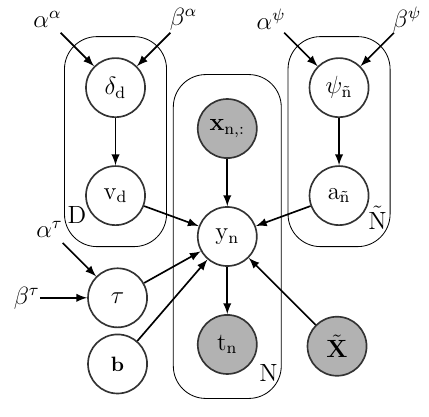}
    \end{minipage} \quad
    \begin{minipage}{.4\textwidth}
        \begin{equation}
            \yn \sim \N(\xn\vlam\Xst\as + b , \tau^{-1})
        \end{equation}
        \begin{equation}
            \an \sim \N(0, \fin^{-1})
        \end{equation}
        \begin{equation}
            \fin \sim \G(\anfi, \bnfi)
        \end{equation}
        \begin{equation}
            \vd \sim \N_{F}(0,\alfad^{-1})
        \end{equation}
        \begin{equation}
            \alfad \sim \G(\adalfa,\bdalfa)
        \end{equation}
        \begin{equation}
            b \sim \N(0,1)
        \end{equation}
        \begin{equation}
            \tau \sim \G(\atau, \btau)
        \end{equation}
        \begin{equation}
            p(t_n|\yn) = e^{\yn t_n}\sigma(-\yn)
        \end{equation}
  \end{minipage}
  \caption{Diagram of the graphical model of RFVM for classification tasks. Grey circles denote observed variables, and white circles unobserved random variables. The nodes without a circle correspond to the hyperparameters. Also, the top-left plate, which factorizes over $D$ represents the FS capability, while the top-right plate, which factorizes over $\tilde{N}$, represents the RVS. The  central plate factorizes over $N$, what allows the model to consider independence between samples.}
  \label{fig:GM_RFVM_class} 
\end{figure}

Figure \ref{fig:GM_RFVM_class} shows the graphical representation of the generative model, including the elements that enable the RVS and the FS.

\subsection{Model Inference}
\label{sec:inference}

The posterior distribution $p(\mathbf{\Theta}| \mathbf{t},\mathbf{X})$, where $\mathbf{\Theta} = \left[\mathbf{y},\mathbf{a}, \mathbf{v}, b, \boldsymbol{\delta}, \boldsymbol{\psi}, \tau\right]$, cannot be obtained in an analytical way since it depends on an intractable integral\footnote{Within the Bayes rule it is defined the integral between the likelihood (which is gaussian in this case) of the model and the variable prior distributions. As some of the variables do not follow a gaussian distribution (gamma and normal folded), the integral cannot be expressed as a convolution of gaussians and subsequently cannot be solved in a closed form.}, we employ variational inference to approximate it by a variational distribution $q(\bm{\Theta})$. This distribution  $q(\bm{\Theta})$ must be a good approximation to the real posterior. To measure the quality of this approximation we can resource to the computation of the Kullback-Leibler (KL) divergence between  $q(\bm{\Theta})$ and the real posterior $p(\mathbf{\Theta}|\mathbf{t},\mathbf{X})$:

\begin{equation}
    KL(q|p) = \int q(\bm{\Theta}) \ln\left( \frac{q(\bm{\Theta})}{p(\bm{\Theta}|\mathbf{t},\mathbf{X})}\right) d\bm{\Theta} = \int q(\bm{\Theta}) \ln\left( \frac{q(\bm{\Theta})}{p(\bm{\Theta},\mathbf{t},\X)}\right) d\bm{\Theta} + \ln p(\mathbf{t},\mathbf{X}).
\end{equation}

The minimization of $KL(q|p)$ can be employed to obtain the optimal  $q(\bm{\Theta})$. Notice that all the influence of $q(\bm{\Theta})$ in the value of $KL(q|p)$ is included in 
\begin{equation}
\label{eq:teta}
    L(q) = -\int q(\bm{\Theta}) \ln\left( \frac{q(\bm{\Theta})}{p(\bm{\Theta},\mathbf{t},\X)}\right) d\bm{\Theta}
\end{equation}

Therefore, minimizing $KL(q|p)$ with respect to 
$q(\bm{\Theta})$ is equivalent to maximize $L(q)$. Equation \eqref{eq:teta} can be further refined to yield
\begin{equation}
\begin{split}
    L(q) & = -\int q(\bm{\Theta}) \ln\left( \frac{q(\bm{\Theta})}{p(\bm{\Theta},\mathbf{t},\X)}\right) d\bm{\Theta} = \int q(\bm{\Theta}) \ln\left(q(\bm{\Theta})\right) 
    - \int q(\bm{\Theta})\ln\left(p(\bm{\Theta}, \mathbf{t}, \X)\right) \\
    & = \E_{q}\left[\ln (q(\bm{\Theta}))\right] - \E_{q}\left[\ln (p(\bm{\Theta}, \mathbf{t}, \X))\right]
    \label{eq:elboo}
\end{split}
\end{equation}
where $\E_{q}[\ln p(\mathbf{t},\X,\bm{\Theta})]$ corresponds to the expectation of the logarithm of the joint distribution and $\E_{q}[\ln 
 q(\bm{\Theta})]$ to the entropy of $q(\mathbf{\Theta})$.

The posterior $q(\bm{\Theta})$ is defined using the mean-field approximation  \citep{blei2017variational}, which assumes that $q(\bm{\Theta})$ can be factorized over all model variables as
\begin{equation}
     q(\boldsymbol\Theta) = \prod q_i(\bm{\Theta}_i) = q_{\mathbf{y}}(\y)q_{\mathbf{a}}\left(\textbf{a}\right)q_{\boldsymbol{\psi}}(\boldsymbol{\psi}) q_{\boldsymbol{\alpha}}\left(\boldsymbol\alpha\right)q_{\tau}(\tau)q_{b}(b)\prod_{d=1}^{D}q_{\vd}(\vd) 
\end{equation}

Note that the $\{\vd\}_{d=1}^D$ are assumed to be independent, each with a separate  $q_{\vd}(\vd)$. This assumption avoids the calculation and construction of a full $D \times D$ covariance matrix for $\mathbf{v}$, which is usually prohibitive in fat-data problems.

Due to the fact that  $q(\mathbf \Theta)$ is a product of the $q_i(\mathbf \Theta_i)$, we can express $L(q)$ as:

\begin{equation}
\begin{split}
    L(q) & = \int q(\bm{\Theta}) \ln\left( \frac{p(\bm{\Theta},\mathbf{t},\X)}{q(\bm{\Theta})}\right) d\bm{\Theta} = \int \prod_{i}q_i\left[\ln(p(\bm{\Theta},\mathbf{t},\X)) - \sum_i \ln(q_i)\right]d\bm{\Theta} \\
    & = \int \prod_i q_i \ln(p(\bm{\Theta},\mathbf{t},\X))d\bm{\Theta} - \int \prod_i q_i \sum_i \ln(q_i)d\bm{\Theta} \\
    & = \int q_j \prod_{i \neq j} q_i \ln(p(\bm{\Theta},\mathbf{t},\X)) d\bm{\Theta} - \int q_j \prod_{i \neq j} q_i \left(\ln(q_j) + \sum_{i \neq j}\ln(q_i)\right)d\bm{\Theta} \\
    & = \int q_j \prod_{i \neq j} q_i \ln(p(\bm{\Theta},\mathbf{t},\X)) d\bm{\Theta} - \int q_j \prod_{i \neq j} q_i \sum_{i \neq j}\ln(q_i) d\bm{\Theta} - \int q_j \prod_{i \neq j} q_i \ln(q_j)d\bm{\Theta}
    \\& = \int q_j \left[\int \prod_{i \neq j} q_i \ln p(\mathbf{t},\X,\bm{\Theta})d\bm{\Theta}_i \right]d\bm{\Theta}_j - \int q_j \ln q_jd\bm{\Theta}_j + const \\
    & = \int q_j\ln(f_j)d\bm{\Theta}_j - \int q_j\ln(q_j)d\bm{\Theta}_j + const.
    \label{eq:MFA}
\end{split}
\end{equation}
where $\ln (f_j) = \mathbb{E}_{-q_j}[\ln p(\textbf{t},\textbf{X},\bm{\Theta})]$, and $\mathbb{E}_{-q_j}$ means the expectation over all the random variables in $q$ except the $j$th one.

 Therefore, to maximize $L(q)$, the model has to find the distribution $q_j$ that best  approximates $f_j$. Hence, the optimal solution has the equation:
\begin{equation} 
\ln q_j^* = \mathbb{E}_{-q_j}[\ln p(\textbf{t},\textbf{X}, \bm{\Theta})] + const. \label{eq:lnqopt}
\end{equation} 

Eq. (\ref{eq:lnqopt}) is applied iteratively for each $q_j$ until convergence. Notice that, at each iteration, the corresponding $q_j$ is updated in order to maximize $L(q)$ while the other $q_i$ ($i \ne j$) remain frozen.
 
\begin{table*}[ht!]
    \begin{adjustbox}{max width=\textwidth}
        \centering
        \renewcommand{\arraystretch}{2}
        \setlength{\tabcolsep}{25pt}
        \begin{tabular}{c c c}
        \hline
        Variable & $q^{*}$ distribution & Parameters\\
        \hline
        $a_{\hat{n}}$ & $\N(a_{\hat{n}}|\langle a_{\hat{n}}\rangle,\Sigma_{\A})$ & \begin{tabular}{@{}c@{}}$\acov^{-1} = \tauE \sumn(\Xs\xlam\vvt\xlam\Xst) + \filamE$ \\ $\aE = \tauE(\langle\ys\rangle^T - \langle\bb\rangle\mathbf{1}^{T}_{N})\X\vlamE\Xst\acov$\end{tabular}\\
        \hline
        $\vd$ & $\N_{F}(v_{d}|\langle \tilde{v}_{d}\rangle,\tilde{\sigma}_{d}^2)$ & \begin{tabular}{@{}c@{}c@{}c@{}c@{}} $\tilde{\sigd}^2 = \vdE^2 + \sigd^2 - \tilde{\vdE}^2$\\ $\tilde{\vdE} = \sqrt{\frac{2}{\sigd^2\pi}}e^{-\frac{\vdE}{2\sigd^2}} + \vdE \left(1 - 2\Phi\left(-\frac{\vdE}{\sigd}\right) \right) $ \\ where\\ $\sigd^2 = \tauE\xsdt\aat\xsd\sumn(\xnd^2) + \langle\alfad\rangle$ \\ $\langle\vd\rangle = (\sigd^{-2})\sumn\left[\tauE(\langle\yn\rangle - \bE)\xnd\xsdt\aE \right.$ \\$\left. - \underset{(d \neq \hat{d})}{\sumdrara}\left[\tauE\langle v_{\hat{d}}\rangle \tilde{\mathbf{x}}_{:,\hat{d}}^T\aat\xsdt\xnd x_{n,\hat{d}}\right]\right]$ \end{tabular}\\
        \hline
        $\psi_{\hat{n}}$ & $\Gamma(\psi_{\hat{n}}|\alpha_{\hat{n}}^{\psi},\beta_{\hat{n}}^{\psi})$ & \begin{tabular}{@{}c@{}}$\afi = \unmed + \acerofi$ \\ $\bfi = \bcerofi + \unmed diag(\aat)$\end{tabular}\\
        \hline
        $\delta_{d}$ & $\Gamma(\alpha_{d}|\alpha_{d}^{\delta},\beta_{d}^{\delta})$ & \begin{tabular}{@{}c@{}}$\aalfa = \unmed + \aceroalfa$ \\ $\balfa = \bceroalfa + \unmed diag(\vvt)$\end{tabular}\\
        \hline
        $b$ & $\N(0,1)$ & \begin{tabular}{@{}c@{}}$\Sigma_{b}^{-1} = N \tauE +1$ \\ $\langle b\rangle = \tauE\left[-\mathbf{1}_{N}\X\vlamE\Xst\aE + \sum_{n=1}^{N}\langle \yn\rangle\right]\Sigma_{b}$\end{tabular}\\
        \hline
        $\taus$ & $\Gamma(\taus|\alpha^{\taus},\beta^{\taus})$ & \begin{tabular}{@{}c@{}}$\atau = \frac{N}{2} + \acerotau$ \\ $\btau = \bcerotau + \unmed\left[\langle\ys^T\ys\rangle -2\langle\ys\rangle^T\X\vlamE\Xst\aE \right.$\\
        $+ 2\langle\bb\rangle\mathbf{1}_{N}\mathbf{X}\vlamE\Xst\aE + N\langle b^2\rangle$\\ $\left.+ \sumn Trace\left(\vvt\xlam\Xst\aat\Xs\xlam\right) -2\langle\ys\rangle^T\langle\bb\rangle\mathbf{1}_{N}^T\right]$\end{tabular}\\
        \hline
        $\yn$ & $\N(y_{n}|\langle y_{n}\rangle,\Sigma_{\y})$ & \begin{tabular}{@{}c@{}}$\Sigma_{\ys}^{-1} = \tauE I +2\Lambda_{\bm{\xi}}$ \\ $\langle \ys\rangle = (\mathbf{t} - \unmed\mathbf{1}_{N} + \tauE\aE\Xs\vlamE \X^\top + \tauE\bE \mathbf{1}_{N}) \Sigma_{\ys}$\end{tabular}\\
        \hline
        \end{tabular}
    \end{adjustbox}
    \caption{$q^{*}$ update rule for all the r.v. in the RFVM model obtained using the mean-field approximation. These expressions were obtained using Eq. (\ref{eq:lnqopt}). Also, $\alpha_0^{\mathbf{z}}$ and $\beta_0^{\mathbf{z}}$ denote the initial values of the parameters of the gamma distribution of a r.v. $\mathbf{z}$. Furthermore, $\Phi$ denotes the cumulative normal distribution, also called probit distribution.}
    \label{tab:my_label}
\end{table*}

Also, to  incorporate Eq. (\ref{eq:t}) in the mean-field approximation we need to approximate p(t|y) with a gaussian. For this purpose we follow the idea presented in \cite{jaakkola2000bayesian} to define a lower bound on the log-likelihood of $p(t_n|\yn)$ based on a first-order Taylor series expansion over $\yn^2$ (see \ref{sec_logbound})
\begin{equation}
    \ln p(t_n = 1|\yn) = 
    e^{\yn}\sigma(-\yn) \geqslant h(\yn, \xi_n) = 
    e^{\yn}\sigma(\xi_n)e^{-\frac{\yn + \xi_n}{2} - \lambda(\xi_n)(\yn^2 - \xi_n^2)},
    \label{eq:lbb}
\end{equation}
where $\lambda(a) = \frac{1}{2a}(\sigma(a) - \unmed)$ and $\xi_n$ is a variational parameter that defines the center of the Taylor expansion  of $\yn^2$. This approach enables to treat $p(t_n|\yn)$  as a gaussian distribution that conjugates with the rest of the distributions of the model, and subsequently allows to obtain an analytical solution by means of the inference procedure explained above.

Deriving Eq. \eqref{eq:lnqopt} we get the mean-field factor update rules depicted in Table \ref{tab:my_label} (mathematical developments detailed in \ref{sec_Ap1}). Note that, as the parameters of the distributions are not independent, the inference process needs to interatively update parameters of the distributions of the variables by maximizing $L(q)$ until convergence. Once all the variables have been updated in the current iteration, the following convergence rule is evaluated: $L(q)_{T-100} > L(q)_{T}(1 - 10^{-8})$, where $L(q)_{T}$ and $L(q)_{T-100}$  represent the $L(q)$ values at iterations $T$ and $T-100$, respectively. To analyze the evolution of $L(q)$ we can rewrite Eq. \eqref{eq:elboo} by using the expressions of Table \ref{tab:my_label} as
\begin{equation}
\begin{split}
    L(q) & = -\unmed\tilde{N}\ln|\acov| - \unmed \sum_{d=1}^{D}\ln|\sigma_{d}^2| - \unmed\ln|\Sigma_{b}| + \left(-\unmed N -\acerotau+2\right)\ln \btau \\ & + \left(-\unmed -\aceroalfa
    +2\right)\sum^{D}_{d = 1}\ln \bdalfa + \left(-\unmed - \acerofi +2\right)\sum^{\tilde{N}}_{\tilde{n} = 1}\ln \bnfi -\unmed\langle b^2\rangle \\ &
    - \frac{N}{2}\ln{|\Sigma_{\mathbf y}|} + \langle \mathbf y \rangle^{\top} \mathbf t  + \sum_{n=1}^{N}\left(\ln(\sigma(\xi_n)) - \frac{1}{2}\left(\yn + \xi_n\right)\right)\\ &
    - \frac{1}{2}(\lambda(\boldsymbol{\xi}^T)(diag(\langle \mathbf y \mathbf y^\top \rangle) - \boldsymbol{\xi}^{\bullet 2})
    \label{eq:ELBO}
\end{split}
\end{equation}
where $\sigma_{d}^2$ is the variance of $\vd$, $\boldsymbol{\xi} = \left[\xi_1,\ldots, \xi_N\right]^\top$ and $\boldsymbol{\xi}^{\bullet 2}$ means square all the elements of the vector $\boldsymbol{\xi}$; also $\alpha_0^{\mathbf{z}}$ and $\beta_0^{\mathbf{z}}$ denote the initial values of the gamma distribution of a certain r.v. $\mathbf{z}$. We also use this expression to optimize $\boldsymbol{\xi}$, just by direct optimization. The complete development of the $L(q)$ terms is detailed in \ref{sec_Ap2}.

Finally, we incorporated a prunning procedure within the training iterative process that removes the components of $\mathbf{a}$ and $\mathbf{v}$ associated to the irrelevant features and samples. To do so, we set a threshold at $10^{-2}$ and $10^{-3}$ times the maximum absolute value of the associated weight vectors ($\langle\bf v\rangle$ and $\langle\bf a\rangle$ depicted in Table \ref{tab:my_label}) for FS and RVS, respectively. Also, to select the best candidates to be RVs, we initialized $\tilde{\X}$ with all the training samples ($\tilde{\X} = \X$).

\subsection{RFVM Posterior Predictive Distribution}
\label{sec:reg_pred2}

As the model is intended to work in binary classification scenarios,  the posterior predictive distribution  of the label of a test observation $\mathbf{x}_{*,:}$ is given by:
\begin{equation}
    p(t^* = 1|\mathbf{t},\X, \mathbf{x}_{*,:}) = \int_{\chi_{\yas}} p(t^* = 1|\yas)p(\yas|\mathbf{t},\X,\mathbf{x_{*,:}})d\yas.
    \label{eq:dist_clas}
\end{equation}
As $p(t^* = 1|\yas) = \sigma(\yas)$, we can express Eq. (\ref{eq:dist_clas}) as
\begin{equation}
\label{eq:true_post}
    p(t^* = 1|\mathbf{t}) = \int_{\chi_{\yas}} \sigma( \yas)p(\yas|\mathbf{t},\X,\mathbf{x_{*,:}})d\yas.
\end{equation}

As the Eq. (\ref{eq:true_post}) can not be solved analytically, we follow \cite{bishop2006pattern}, that suggests the approximation of $\sigma(\yas)$ with a gaussian cumulative distribution function $\Phi(\kappa \yas)$, where $\kappa$ is a parameter that rescales the horizontal axis of the probit function to match a sigmoid that transforms $\yas$ into the range (0,1). Now, the approximated integral can be obtained in a closed form as
\begin{equation}
    p(t^* = 1|\mathbf{t},\yas) \simeq \int_{\chi_{\yas}} \Phi(\kappa \yas)p(\yas|\mathbf{t},\X,\mathbf{x_{*,:}})d\yas = \Phi\left(\frac{\langle \yas\rangle}{(\kappa^{-2} +\zeta^2_{\yas})^{\frac{1}{2}}}\right),
\end{equation}
where $\langle \yas\rangle$ and $\zeta^2_{\yas}$ represent the mean and variance of $\yas$. Also, \citep{bishop2006pattern} proved that the similarity between the sigmoid and the probit function is maximal when $\kappa^2 = \frac{\pi}{8}$. This yields the final posterior distribution as
\begin{equation}
    p(t^* = 1|\mathbf{t},\yas) \simeq \sigma\left(\frac{\langle \yas\rangle }{(1 + \frac{\pi}{8}\zeta^2_{\yas})^{\unmed}}\right).
\end{equation}

The computation of $\langle \yas\rangle$ and $\zeta^2_{\yas}$ starts by obtaining $p(\yas|\mathbf{t},\X,\mathbf{x}_{*,:})$ for a new sample $\mathbf x_{*,:}$ 

\begin{equation}
    p(\yas|\mathbf{t},\X,\mathbf{x}_{*,:}) = \int p(\yas|\mathbf{t},\X,\mathbf{x}_{*,:}, \bm{\Theta})p(\bm{\Theta}|\mathbf{t},\X) d\bm{\Theta},
\end{equation}
where $\mathbf{\Theta}$ comprises $\A$, $\mathbf{v}$, $\bb$ and $\taus$. Then the true posterior $p(\bm{\Theta}|\mathbf{t},\X)$ is replaced by its approximation $q(\bm{\Theta}) = q_{\mathbf{a}}\left(\textbf{a}\right)q_{\tau}(\tau)q_{b}(\bb)\prod_d^{D}q(\vd)$ to obtain
\begin{equation}
    p(\yas|\mathbf{t},\X,\mathbf{x}_{*,:}) = \int_{\chi_{\Aa}} \int_{\chi_{\vd}} \int_{\chi_{\bb}} \int_{\chi_{\taus}} p(\yas|\mathbf{t},\X,\mathbf{x}_{*,:}, \boldsymbol{\Theta})q_{\mathbf{a}}\left(\textbf{a}\right)q_{\tau}(\tau)q_{b}(\bb)\prod_d^{D}q_{\vd}(\vd)d\mathbf{a} d \vd d b d\tau.
\end{equation}

Since the integral over the noise $\taus$ is intractable, we perform a point estimation around its posterior mean, and approximate the predictive distribution as:
\begin{equation}
    p(\yas|\mathbf{t},\X,\mathbf{x}_{*,:}) \simeq \int_{\chi_{\Aa}} \int_{\chi_{\vd}} \int_{\chi_{\bb}} p(\yas|\mathbf{t},\X,\mathbf{x}_{*,:}, \boldsymbol{\Theta}_{-\taus},\tauE)q_{\mathbf{a}}\left(\textbf{a}\right)\prod_d^{D}q_{\vd}(\vd)q_{b}(b)d\mathbf{a} d\vd d\mathbf{b}.
\end{equation}

After, since $\mathbf a$ and $\mathbf b$ follow Gaussian distributions, we can marginalize them as:
\begin{equation}
\begin{split}
    p(\yas|\mathbf{t},\X,\mathbf{x}_{*,:}) \simeq \int_{\chi_{\vd}} \left[ \N(\mathbf{x}_{*,:}\vlam\Xst\aE + \langle b\rangle, \tauE^{-1} + \right. 
    \\ \left. \mathbf{x}_{*,:}\vlam\Xst\Sigma_{\A}\Xs\vlam\mathbf{x}_{*,:}^T + \Sigma_{b})\prod_d^{D}q_{\vd}(\vd)d\vd\right].
    \label{eq:pred_fin_2}
\end{split}
\end{equation}

Next, since $q(\mathbf{v})$ is a normal folded distribution, no analytical method is available to marginalize over $\mathbf{v}$. However, as the solution of the integral of Eq. (\ref{eq:pred_fin_2}) yields a unimodal non-symmetric distribution \cite{bishop2006pattern} we can finally compute its mean and variance as:

\begin{align}
    \langle y_* \rangle  &= \mathbf{x_{*,:}}\vlamE\Xst\aE + \langle b\rangle,
    \label{eq:mean_pred}\\
    \zeta^2_{y_*} &= \mathbf{x}_{*,:}\vlamE\Xst\Sigma_{\A}\Xs\vlamE\mathbf{x}_{*,:}^\top + \tauE + \Sigma_{b} + \sum_d^{D} \left(x_{*,d}^2 \xsdt\sigma_{\vd}\langle \A^\top\A\rangle\xsd \right).
    \label{eq:cov_pred}
\end{align}

Further mathematical details are available in \ref{sec_Ap3}.

\section{Evaluation of the model in different health applications}
\label{sec:Results}

This section presents an empirical evaluation of the proposed model within different fat-data medical scenarios. It starts with a brief overview of the medical databases under study, the presentation of the baseline models and of the experimental setup. Afterwards, the performance of the RFVM is analyzed across the selected databases in terms of its accuracy, as well as of its FS and RVS capabilities. Next,  the model interpretability is discussed in a well-known cancer database by analyzing in depth the subsets of selected proteins. The section ends with an  evaluation of RFVM computational complexity.

\subsection{Medical databases}

This study uses eight public medical databases characterized by the presence of more features than samples\footnote{The datasets along with some interesting links are available \href{https://jundongl.github.io/scikit-feature/datasets.html}{here}}. All the datasets present a binary classification task. Seven out of the eight databases are related to gene expression data of different cancers (colorectal, lymphoma, protatic, leukemia, lung and glioma). For these datasets, observations correspond to individual cells; each input feature is related to the activation of a certain protein available in the cell genome, i.e., data of gene expression collected from oligonucleotide microarrays, while the labels indicate if the cell belongs to a cancerous tissue or not. Originally multiclass datasets lymphoma and GLI\_85 were transformed into  binary classification tasks consisting in to distinguish between healthy and cancerous tissues. Furthermore, we have the Dorothea dataset, in which each observation corresponds to a chemical drug related with blood coagulation (small organic molecules) and each feature represents different chemical compounds of the molecular structures. Also, the labels indicate if the molecule is binded or not with the thrombin protein (active or inactive). This database  originally includes 100000 features; but, in order to manage feasible training times and reduce the model complexities of the methods that are not prepared to work with fat-data, we have used only the first 8000 features.  Table \ref{tab:databases} summarizes the main characteristics of the 8 databases.

\begin{table}[ht!]
    \centering
    \begin{tabular}{c c c c c c}
    \hline
         Database  & Domain & $\%$ class 1 & Samples & Features & N/D \\
         \hline
          Dorothea & Coagulation Drugs  &90.26  & 1150  & 8000 & 0.143 \\
          colon & Colorectal cancer & 35.43 & 62  & 2000 & 0.031 \\
          lymphoma & Lypmhoblastic cancer  & 52.08 & 96  & 4026 & 0.023 \\
          Prostate$\_$GE & Prostate cancer  & 49.02& 102  & 5966 & 0.017 \\
          leukemia & Leukemic cancer  & 34.72& 72  & 7070 & 0.010 \\
          ALLAML& Leukemic cancer & 65.27& 72   & 7129 & 0.010 \\
          SMK$\_$CAN$\_$187& Lung cancer   & 48.12 & 187  & 19993 & 0.009 \\
          GLI$\_$85 & Glioma tumor  & 30.58& 85 & 22283 & 0.004 \\
         \hline
    \end{tabular}
    \caption{Summary of main characteristics of the databases used in this work sorted by its ratio sample/feature ($N/D$). $\%$ class 1 indicates the $\%$ of data belonging to class 1.}
    \label{tab:databases}
\end{table}

\subsection{Baselines and experimental setup}
\label{sec:Baselines}

To give insights about the significance of integrating FS and sparsity in the RVs when dealing with high-dimensional health-related problems, the following state-of-the-art methods have been selected to serve as baselines in a  comparative analysis. Those baselines are methods that perform either FS, or dual variable selection; simultaneous selection (both FS and dual variable selection)  or no selection at all.

The models that enforce no selection or selection in just one group of variables (primal or dual) include:
\begin{enumerate}
    \item Random Forest (RF) \cite{breiman2001random} where the following hyperparameters were cross-validated: maximum depth within the range $[10, 20, 30, 40, 50]$,  maximum number of features when looking for the best split within the range $[\log D, \sqrt{D}]$,  maximum number of leaves selected among the range $[2^1,\ldots , 2^5]$,  minimum number of samples per split explored within $[2, 5, 10, 20, 50]$ and  total number of estimators among $[200, 400, 600, 800, 1000]$. With respect to sparsity of the solution, RF performs an implicit FS, and features that are not used in any tree in the forest will be removed.
    \item  MultiLayer Perceptron (MLP) \cite{ramchoun2016multilayer} with Adam solver. The best model per simulation was selected by cross-validation from a pool of  4 different hidden layers configurations: (100,50), (100,50,100), (100,  50, 50, 100) and (100, 50, 25, 50, 100). Each element in the tuple means the number of neurons in the corresponding hidden layer. Note that this model does not enforce sparsity in any group of variables.
    \item Logistic Regression  with an elastic-net penalty (LR-elastic) \cite{algamal2015high}. The value for the regularization parameter C was cross validated within the range of $10^{-4}$ to 1000, and the value of the $\ell$1-ratio was cross validated using five values  between 0 and 1. This method performs a FS by means of the $\ell$1 regularization.
    \item Sparse Relevance Vector Machine (SRVM), which performs an implicit sparsity in the dual variables. Due to the probabilistic nature of this model, its hyperparameters were adjusted by maximizing the marginal likelihood.
    \item Sparse Gaussian Process with Automatic Relevance Determination (SGP+ARD), with the number of inducing points cross validated in a  range from 10$\%$ to 100$\%$ of the size of the training set. As a probabilistic method, the rest of its hyperparameters were adjusted by maximizing the marginal likelihood. This method performs a sparsity in the dual variables as its architecture is defined in terms of the inducing points.
\end{enumerate}

The  models that perform simultaneous selection in the primal and dual variables include:
\begin{enumerate}
    \item Support Vector Machine  with an $\ell$1 penalty  (SVM-$\ell$1), with the value of the regularization parameter $C$ cross validated within the same range of values used for LR-elastic. 
    \item Probabilistic Feature Selection and Classification Vector Machine (PFCVM). Since it is a probabilistic method,  its parameters ($\mathbf{v}$ and $\mathbf{a}$) were adjusted by maximizing its likelihood. 

    \item A deterministic version of RFVM named Cross-Optimized LOsses Network (XOLON). XOLON implements the following model:
\begin{equation}
    p(y=1|x) \approx \sigma(f(\mathbf x)) = \sigma(\mathbf x^\top\Lambda_{\mathbf{v}^2}\X^\top\as + b ),
    \label{eq:xolon_class}
\end{equation}
where $\sigma$ is the sigmoid function and $\mathbf v$ and $\mathbf a$ are the parameters to optimise by minimizing the binary cross entropy between model predictions and true targets with a double $\ell_1$ norm regularization over $\mathbf v$ and $\mathbf a$ as
\begin{equation}
    \min \left[ -\frac{1}{N}\sum_{n=1}^{N}\left(\yn\ln\left(\sigma(f(\mathbf x))\right) + (1-\yn)\ln\left(1-\sigma(f(\mathbf x))\right)\right)\right] + \gamma_{v} \sum_{d=1}^D|\vd| +
    \gamma_{a} \sum_{n=1}^N|a_n|,
\end{equation} This resulting regularized loss is iteratively minimized using  Adam \cite{kingma2014adam} with a learning rate cross validated in the range [0.2, 0.1, 0.01, 0.001] XOLON hyperparamenters  $\gamma_{v}$ and $\gamma_{a}$ were cross validated in a range of seven values between -5 and 1 in a logarithmic scale.

\end{enumerate}

All the cross validations involved 5-fold processes, except for the XOLON ones, that were simplified  to 3-fold  processes due to their computational burden. The metric employed to assess performance was the classification accuracy.

Finally, in order to speed up the training process of XOLON and PFCVM, we incorporated the same prunning procedure as RFVM (see Subsection \ref{sec:inference}).

A Python 3.8 implementation of RFVM is available  \href{https://github.com/albello1/RFVM.git}{here}.

\subsection{Performance evaluation}
\label{sec:Reg_res}

To analyze the diagnostic capability of the different methods under consideration, we begin our study by comparing their accuracy in binary classification tasks consisting in to separate samples taken from healthy subjects from those taken from patients, as well as their ability to detect a reduced subset of relevant characteristics (FS) and subjects (RVS). The results are presented in two separate tables. Table \ref{tab:results_standard} compares our model with  baselines that perform single or no selection and Table \ref{tab:results_FSVS} with the  baselines that perform simultaneous selection. The results for each database are organized into three rows, showing classification accuracy, the percentage of selected features, and the size of the set of input observations needed to define the architecture of the model expressed as a percentage of the training set size, respectively. The last  group of three rows of each table presents averaged results, providing a comprehensive overview of each model performance. 

\begin{table}[H]

\scriptsize
\centering
\begin{adjustbox}{max width=\textwidth}
    \begin{tabular}{c c c c c c c}
    \toprule
    ~& RF & SGP+ARD &  MLP & LR-elastic & SRVM & RFVM\\ \midrule
    \multirow{3}{*}{Dorothea}     
                &  \textbf{0.93} $\pm$ \textbf{0.02} & 0.91 $\pm$ 0.02 & 0.89 $\pm$ 0.01 & \textbf{0.93} $\pm$ \textbf{0.01} & 0.86 $\pm$ 0.03 & \textbf{0.93} $\pm$ \textbf{0.05}\\
                & \cellcolor{gray!10} 57.25 $\%$ $\pm$ 6.15 & \cellcolor{gray!10}  100 $\%$  &\cellcolor{gray!10} 100 $\%$ & \cellcolor{gray!10} 0.36 $\%$   $\pm$ 0.04 & \cellcolor{gray!10} 100 $\%$ & \cellcolor{gray!10} \textbf{0.25 $\%$ } $\pm$ \textbf{0.40}  \\
                &  \cellcolor{gray!30} 100 $\%$ & \cellcolor{gray!30}  26.0 $\%$  $\pm$ 8.0 &\cellcolor{gray!30} 100 $\%$ & \cellcolor{gray!30} 100 $\%$ & \cellcolor{gray!30} 1.80 $\%$  $\pm$ 2.47 &\cellcolor{gray!30} \textbf{0.78 $\%$ } $\pm$ \textbf{0.85}  \\     
    \midrule
    \multirow{3}{*}{Colon}     
                &  0.77 $\pm$ 0.08 & 0.64 $\pm$ 0.11 & 0.74 $\pm$ 0.13 & \textbf{0.78} $\pm$ \textbf{0.05} & 0.74 $\pm$ 0.13 &\textbf{0.78} $\pm$ \textbf{0.09} \\
                &  \cellcolor{gray!10} 36.76 $\%$  $\pm$ 5.41 &\cellcolor{gray!10}  100 $\%$  & \cellcolor{gray!10} 100 $\%$ & \cellcolor{gray!10}  1.72 $\%$  $\pm$ 0.64 & \cellcolor{gray!10} 100 $\%$ &\cellcolor{gray!10}  \textbf{0.82 $\%$ } $\pm$ \textbf{0.93}  \\
                & \cellcolor{gray!30} 100 $\%$ &\cellcolor{gray!30}  42.0 $\%$ $\pm$ 31.87 & \cellcolor{gray!30} 100 $\%$ & \cellcolor{gray!30} 100 $\%$ & \cellcolor{gray!30} 14.50 $\%$  $\pm$ 1.41 & \cellcolor{gray!30}\textbf{5.23 $\%$ } $\pm$ \textbf{1.59}  \\     
    \midrule
    \multirow{3}{*}{lymphoma}     
                & 0.88 $\pm$ 0.03 & 0.85 $\pm$ 0.07 & \textbf{0.90} $\pm$ \textbf{0.05} & 0.86 $\pm$ 0.07 & 0.89 $\pm$ 0.06  &\textbf{0.90} $\pm$ \textbf{0.09} \\
                & \cellcolor{gray!10} 15.88 $\%$  $\pm$ 5.34 & \cellcolor{gray!10} 100 $\%$  &\cellcolor{gray!10} 100 $\%$ & \cellcolor{gray!10}  55.43 $\%$  $\pm$ 45.55 & \cellcolor{gray!10} 100 $\%$ & \cellcolor{gray!10} \textbf{3.74 $\%$ } $\pm$ \textbf{6.74}  \\
                & \cellcolor{gray!30} 100 $\%$ &\cellcolor{gray!30} 74.0 $\%$  $\pm$ 25.76 & \cellcolor{gray!30} 100 $\%$ & \cellcolor{gray!30} 100 $\%$ & \cellcolor{gray!30} 10.18 $\%$  $\pm$ 1.69 & \cellcolor{gray!30} \textbf{2.33 $\%$ } $\pm$ \textbf{0.96}  \\     
    \midrule
    \multirow{3}{*}{Prostate$\_$GE}     
                & 0.90 $\pm$ 0.09 & 0.92 $\pm$ 0.05 & 0.89 $\pm$ 0.06 & 0.93 $\pm$ 0.05 &0.57 $\pm$ 0.10 & \textbf{0.94} $\pm$ \textbf{0.07}\\
                & \cellcolor{gray!10} 20.46 $\%$  $\pm$ 8.26 & \cellcolor{gray!10}  100 $\%$  &\cellcolor{gray!10} 100 $\%$ & \cellcolor{gray!10} 28.86 $\%$ $\%$  $\pm$ 36.79 & \cellcolor{gray!10} 100 $\%$ &  \cellcolor{gray!10} \textbf{0.12 $\%$ }  $\pm$ \textbf{0.03} \\
                & \cellcolor{gray!30} 100 $\%$ &\cellcolor{gray!30} 50.0 $\%$  $\pm$ 38.47 & \cellcolor{gray!30} 100 $\%$ & \cellcolor{gray!30} 100 $\%$ & \cellcolor{gray!30} \textbf{1.22 $\%$ } $\pm$ \textbf{0.01} &\cellcolor{gray!30} 1.83 $\%$  $\pm$ 0.62  \\         
    \midrule
    \multirow{3}{*}{leukemia}     
                & \textbf{0.97} $\pm$ \textbf{0.06} & 0.91 $\pm$ 0.05 & 0.90 $\pm$ 0.07 & 0.94 $\pm$ 0.05 & 0.84 $\pm$ 0.10 & \textbf{0.97} $\pm$ \textbf{0.08}\\
                & \cellcolor{gray!10} 11.84 $\%$  $\pm$ 4.29 & \cellcolor{gray!10}  100 $\%$  &\cellcolor{gray!10} 100 $\%$ & \cellcolor{gray!10} 1.98 $\%$  $\pm$ 0.67& \cellcolor{gray!10} 100 $\%$ & \cellcolor{gray!10} \textbf{0.24 $\%$ } $\pm$ \textbf{0.07}  \\
                & \cellcolor{gray!30} 100 $\%$ &\cellcolor{gray!30} 42.0 $\%$  $\pm$ 27.12 & \cellcolor{gray!30} 100 $\%$ & \cellcolor{gray!30} 100 $\%$ & \cellcolor{gray!30} 8.68 $\%$  $\pm$ 1.14 &\cellcolor{gray!30} \textbf{4.85 $\%$ } $\pm$ \textbf{0.67} \\        
    \midrule
    \multirow{3}{*}{ALLAML}     
                & \textbf{0.95} $\pm$ \textbf{0.05} & 0.90 $\pm$ 0.09 & 0.90 $\pm$ 0.08 &\textbf{0.95} $\pm$ \textbf{0.03} & 0.88 $\pm$ 0.06 & 0.94 $\pm$ 0.03 \\
                & \cellcolor{gray!10} 8.37 $\%$  $\pm$ 1.67 & \cellcolor{gray!10}  100 $\%$  &\cellcolor{gray!10} 100 $\%$ & \cellcolor{gray!10} 3.47 $\%$   $\pm$ 1.25 & \cellcolor{gray!10} 100 $\%$ &  \cellcolor{gray!10} \textbf{3.08} $\%$  $\pm$  \textbf{4.6} \\
                & \cellcolor{gray!30} 100 $\%$ &\cellcolor{gray!30} 72.0 $\%$  $\pm$ 26.38 & \cellcolor{gray!30} 100 $\%$ & \cellcolor{gray!30} 100 $\%$ & \cellcolor{gray!30} 8.03 $\%$  $\pm$ 1.09 &\cellcolor{gray!30} \textbf{3.82 $\%$ } $\pm$ \textbf{1.73}  \\   
    \midrule
    \multirow{3}{*}{SMK$\_$CAN$\_$187}     
                & 0.71 $\pm$ 0.06 & 0.62 $\pm$ 0.11 & 0.70 $\pm$ 0.06 & 0.71 $\pm$ 0.05 & 0.61 $\pm$ 0.09 &\textbf{0.73} $\pm$ \textbf{0.07}\\
                & \cellcolor{gray!10} 12.02 $\%$  $\pm$  5.24 & \cellcolor{gray!10}  100 $\%$  & \cellcolor{gray!10} 100 $\%$ & \cellcolor{gray!10} 57.46 $\%$  $\pm$ 41.90 & \cellcolor{gray!10} 100 $\%$ & \cellcolor{gray!10} \textbf{0.11 $\%$ } $\pm$ \textbf{0.03}   \\
                & \cellcolor{gray!30} 100 $\%$ & \cellcolor{gray!30} 30.0 $\%$ $\pm$ 21.91 &\cellcolor{gray!30} 100 $\%$ & \cellcolor{gray!30} 100 $\%$ & \cellcolor{gray!30} \textbf{0.67 $\%$ } $\pm$ \textbf{0.23} &\cellcolor{gray!30} 2.80 $\%$  $\pm$ 1.28  \\   
    \midrule
    \multirow{3}{*}{GLI$\_$85}     
                & \textbf{0.85} $\pm$ \textbf{0.05} & 0.83 $\pm$ 0.09 & 0.79 $\pm$ 0.12 & 0.84 $\pm$ 0.11 & 0.71 $\pm$ 0.04 & 0.80 $\pm$ 0.04 \\
                & \cellcolor{gray!10} 4.68 $\%$  $\pm$ 0.94 & \cellcolor{gray!10} 100 $\%$ & \cellcolor{gray!10} 100 $\%$ & \cellcolor{gray!10} 1.41 $\%$  $\pm$ 0.32& \cellcolor{gray!10} 100 $\%$ & \cellcolor{gray!10} \textbf{1.09 $\%$ } $\pm$ \textbf{1.29} \\
                & \cellcolor{gray!30}  100 $\%$ & \cellcolor{gray!30} 44.0 $\%$  $\pm$ 20.59 & \cellcolor{gray!30} 100 $\%$ & \cellcolor{gray!30} 100 $\%$ & \cellcolor{gray!30} \textbf{1.47 $\%$ } $\pm$ \textbf{0.0} &\cellcolor{gray!30} 3.66 $\%$  $\pm$ 0.74 \\  
    \midrule
    \midrule
    \multirow{3}{*}{Average}     
                & \textbf{0.87} $\pm$ \textbf{0.05}  & 0.82 $\pm$ 0.07 &  0.83 $\pm$ 0.07 & 0.86 $\pm$ 0.05 & 0.76  $\pm$ 0.07 & \textbf{0.87} $\pm$  \textbf{0.06}  \\
                & \cellcolor{gray!10} 20.91 $\%$  $\pm$ 4.66 & \cellcolor{gray!10} 100 $\%$  &\cellcolor{gray!10} 100 $\%$ & \cellcolor{gray!10} 18.84 $\%$  $\pm$ 22.81 & \cellcolor{gray!10}  100 $\%$  & \cellcolor{gray!10} \textbf{1.18 $\%$ } $\pm$ \textbf{1.76} \\
                & \cellcolor{gray!30} 100 $\%$ &\cellcolor{gray!30} 47.5 $\%$  $\pm$ 25.01  & \cellcolor{gray!30} 100 $\%$ & \cellcolor{gray!30} 100 $\%$ & \cellcolor{gray!30} 5.81 $\%$  $\pm$ 1.01 & \cellcolor{gray!30} \textbf{3.16 $\%$ } $\pm$ \textbf{1.05} \\ 
    \bottomrule
    \end{tabular}
    \end{adjustbox}
    \caption{Results (mean and standard deviation of a 5-fold cross validation) of the proposed RFVM and the baselines with single or no sparsity. Results include the classification accuracy (white), the percentage of  features found relevant (light gray), and the percentage of  training observations needed to define the classification function  (dark gray).}
    \label{tab:results_standard}
\end{table}

If we first analyze the results in Table \ref{tab:results_standard}, we can appreciate that RFVM either matches or outperforms the best baseline in 6 out of 8 databases.
Among the baselines, RF gives the best results, matching the performance of RFVM in 2 databases and surpassing it in $GLI\_85$ and $ALLAML$. However, as some deviation intervals overlap, the accuracy results do not seem to provide a significant improvement over the state-of-the-art. Regarding FS, our proposed model achieves a significantly more compact selection. In some of the problems the number of features selected by RFVM is 3 orders of magnitude smaller than those of RF and LR-elastic. In a clinical context, the use of RFVM as core component for a diagnosis system will potentially save a lot of resources in terms of medical tests to arrive at similar results in terms of diagnostic accuracy. Regarding sparsity in the dual variables, RFVM models are defined in terms of a significantly smaller number of input observations than SRVM or SGP+ARD. It is remarkable how the combination of both sparsities and selections in a same formulation helps RFVM recruit a number of RVs smaller than that of SRVM in practically all the datasets. From the medical point of view, it seems that RFVM is capable of delivering the identification of a compact subset of critical subjects for the characterization of the disease under study.

\begin{table}[H]

\scriptsize
\centering
\begin{tabular}{c c c c c}
    
\toprule
~ & SVM-$\ell$1 & Xolon & PFCVM & RFVM\\ \midrule

\multirow{3}{*}{Dorothea}     
            &   \textbf{0.93} $\pm$ \textbf{0.01} & 0.89 $\pm$ 0.05 & 0.84 $\pm$ 0.01 & \textbf{0.93} $\pm$ \textbf{0.05}\\
            &  \cellcolor{gray!10} 1.16 $\%$  $\pm$ 1.31 & \cellcolor{gray!10} 48.25 $\%$ $\pm$ 16.01 & \cellcolor{gray!10} 73.77 $\%$ $\pm$ 30.99 & \cellcolor{gray!10} \textbf{0.25 $\%$ } $\pm$ \textbf{0.40}  \\
            &   \cellcolor{gray!30} 87.60 $\%$  $\pm$ 24.45 & \cellcolor{gray!30} 79.39 $\%$ $\pm$ 30.45 & \cellcolor{gray!30} 99.56 $\%$ $\pm$ 0.81 & \cellcolor{gray!30} \textbf{0.78 $\%$ } $\pm$ \textbf{0.85}  \\     
\midrule
\multirow{3}{*}{Colon}     
            &  0.77 $\pm$ 0.07 & 0.71 $\pm$ 0.11 & 0.77 $\pm$ 0.08 & \textbf{0.78} $\pm$ \textbf{0.09} \\
            & \cellcolor{gray!10} 1.67 $\%$  $\pm$ 0.93 &  \cellcolor{gray!10} 89.91 $\%$  $\pm$ 3.74 & \cellcolor{gray!10} 62.27 $\%$  $\pm$ 5.92 & \cellcolor{gray!10}  \textbf{0.82 $\%$ } $\pm$ \textbf{0.93}  \\
            &  \cellcolor{gray!30} 82.61 $\%$  $\pm$ 16.02 & \cellcolor{gray!30} 82.44 $\%$  $\pm$ 24.13 & \cellcolor{gray!30} 98.39 $\%$  $\pm$ 0.80 & \cellcolor{gray!30} \textbf{5.23 $\%$ } $\pm$ \textbf{1.59}  \\     
\midrule
\multirow{3}{*}{lymphoma}     
            &  \textbf{0.90} $\pm$ \textbf{0.02} & 0.73 $\pm$ 0.16 & \textbf{0.90} $\pm$ \textbf{0.05} & \textbf{0.90} $\pm$ \textbf{0.09} \\
            & \cellcolor{gray!10} \textbf{0.86 $\%$ } $\pm$ \textbf{0.15} &  \cellcolor{gray!10} 83.18 $\%$  $\pm$ 17.23 & \cellcolor{gray!10} 76.15 $\%$  $\pm$ 15.44 & \cellcolor{gray!10} 3.74 $\%$  $\pm$ 6.74  \\
            &   \cellcolor{gray!30} 61.80 $\%$  $\pm$ 12.62 & \cellcolor{gray!30} 83.63 $\%$  $\pm$ 18.21 & \cellcolor{gray!30} 97.18 $\%$  $\pm$ 3.63 & \cellcolor{gray!30} \textbf{2.33 $\%$ } $\pm$ \textbf{0.96}  \\     
\midrule
\multirow{3}{*}{Prostate$\_$GE}     
            &  0.93 $\pm$ 0.06 & 0.82 $\pm$ 0.03 & 0.72 $\pm$ 0.16 & \textbf{0.94} $\pm$ \textbf{0.07}\\
            &  \cellcolor{gray!10} 0.75 $\%$  $\pm$ 0.25 & \cellcolor{gray!10} 51.09 $\%$  $\pm$ 7.85 & \cellcolor{gray!10} 45.04 $\%$  $\pm$ 8.43 & \cellcolor{gray!10} \textbf{0.12 $\%$ }  $\pm$ \textbf{0.03} \\
            &  \cellcolor{gray!30} 55.17 $\%$  $\pm$ 13.71 &  \cellcolor{gray!30} 93.15 $\%$  $\pm$ 6.75 & \cellcolor{gray!30} 96.81 $\%$  $\pm$ 2.12 &\cellcolor{gray!30} \textbf{1.83 $\%$ } $\pm$ \textbf{0.62}  \\         
\midrule
\multirow{3}{*}{leukemia}     
            & \textbf{0.97} $\pm$ \textbf{0.03} & 0.74 $\pm$ 0.17 & 0.83 $\pm$ 0.12 & \textbf{0.97} $\pm$ \textbf{0.08}\\
            &   \cellcolor{gray!10} 0.47 $\%$  $\pm$ 0.12 & \cellcolor{gray!10} 57.62 $\%$  $\pm$ 13.41 & \cellcolor{gray!10} 89.01 $\%$  $\pm$ 21.99 & \cellcolor{gray!10} \textbf{0.24 $\%$ } $\pm$ \textbf{0.07}  \\
            & \cellcolor{gray!30} 73.94 $\%$  $\pm$ 12.31 & \cellcolor{gray!30} 50.19 $\%$  $\pm$ 29.57 & \cellcolor{gray!30} 100.0 $\%$  $\pm$ 0.0 &\cellcolor{gray!30} \textbf{4.85 $\%$ } $\pm$ \textbf{0.67} \\        
\midrule
\multirow{3}{*}{ALLAML}     
            &  \textbf{0.95} $\pm$ \textbf{0.05} & 0.92 $\pm$ 0.09 & 0.90 $\pm$ 0.07 & 0.94 $\pm$ 0.03 \\
            &  \cellcolor{gray!10} \textbf{0.46 $\%$ } $\pm$ \textbf{0.11} & \cellcolor{gray!10} 65.77 $\%$  $\pm$ 13.73 & \cellcolor{gray!10} 92.86 $\%$  $\pm$ 14.27 &  \cellcolor{gray!10} 3.08 $\%$  $\pm$  4.63 \\
            &  \cellcolor{gray!30} 65.66 $\%$  $\pm$ 11.01 & \cellcolor{gray!30} 75.77 $\%$  $\pm$ 36.74 & \cellcolor{gray!30} 100.0 $\%$  $\pm$ 0.0 &\cellcolor{gray!30} \textbf{3.82 $\%$ } $\pm$ \textbf{1.73}  \\   
\midrule
\multirow{3}{*}{SMK$\_$CAN$\_$187}     
            &  0.70 $\pm$ 0.03 & 0.53 $\pm$ 0.12 & 0.57 $\pm$ 0.09 & \textbf{0.73} $\pm$ \textbf{0.07}\\
            & \cellcolor{gray!10} 0.78 $\%$  $\pm$ 0.58 & \cellcolor{gray!10}  99.89 $\%$  $\pm$ 0.01  & \cellcolor{gray!10} 48.98 $\%$  $\pm$ 8.61 & \cellcolor{gray!10} \textbf{0.11 $\%$ } $\pm$ \textbf{0.03}   \\
            &  \cellcolor{gray!30} 68.98 $\%$  $\pm$ 5.90 & \cellcolor{gray!30} 100.0 $\%$  $\pm$ 0.0 & \cellcolor{gray!30} 97.72 $\%$  $\pm$ 0.91 & \cellcolor{gray!30} \textbf{2.80 $\%$ } $\pm$ \textbf{1.28}  \\   
\midrule
\multirow{3}{*}{GLI$\_$85}     
            &  \textbf{0.85} $\pm$ \textbf{0.06} & 0.82  $\pm$ 0.07 & 0.76 $\pm$ 0.15 & 0.80 $\pm$ 0.04 \\
            & \cellcolor{gray!10} \textbf{0.22 $\%$ } $\pm$ \textbf{0.05} & \cellcolor{gray!10} 62.09 $\%$  $\pm$ 5.39 & \cellcolor{gray!10} 38.92 $\%$  $\pm$ 3.12 & \cellcolor{gray!10} 1.09 $\%$  $\pm$ 1.29 \\
            &  \cellcolor{gray!30} 84.70 $\%$  $\pm$ 7.41 & \cellcolor{gray!30} 98.52 $\%$  $\pm$ 1.86 & \cellcolor{gray!30} 98.52 $\%$  $\pm$ 1.86 & \cellcolor{gray!30} \textbf{3.66 $\%$ } $\pm$ \textbf{0.74} \\  
\midrule
\midrule

\multirow{3}{*}{Average}     
            &   \textbf{0.87} $\pm$ \textbf{0.04}  & 0.77 $\pm$ 0.10 & 0.78 $\pm$ 0.09 & \textbf{0.87} $\pm$ \textbf{0.07} \\
            &  \cellcolor{gray!10} \textbf{0.79 $\%$ } $\pm$ \textbf{0.43} & \cellcolor{gray!10} 69.72 $\%$ $\pm$ 9.67 &\cellcolor{gray!10} 65.87 $\%$ $\pm$ 13.59 & \cellcolor{gray!10} 1.58 $\%$   $\pm$  1.34 \\
            &  \cellcolor{gray!30} 72.55 $\%$  $\pm$ 12.92 &\cellcolor{gray!30} 82.88 $\%$  $\pm$ 18.46 & \cellcolor{gray!30} 98.52 $\%$ $\pm$ 1.26 & \cellcolor{gray!30} \textbf{3.16 $\%$ } $\pm$ \textbf{1.05} \\ 
\bottomrule
\end{tabular}
\caption{Results (mean and standard deviation of a 5-fold cross validation) corresponding to models with simultaneous sparsity in the primal and dual variables. Results include the classification accuracy (white), the percentage of  features found relevant (light gray), and the percentage of  training observations needed to define the classification function  (dark gray).}
\label{tab:results_FSVS}
\end{table}

On the other hand, Table \ref{tab:results_FSVS} shows that, in terms of accuracy, RFVM clearly outperforms both XOLON and PFCVM and closely competes with SVM-$\ell$1 (both achieve  similar accuracies in average). In particular, both models get similar performance in 6 databases, while the RFVM clearly outperforms in average the SVM-$\ell$1 in $SMK\_CAN\_187$ and SVM-$\ell$1 outperforms in average in $GLI\_85$. Also, if we analyze the results as a whole, due to the overlapping of the deviation intervals, the accuracy results do not provide relevant improvements in comparison with the SVM-$\ell$1 model. These results also point out how the Bayesian formulation of RFVM is specially suited for fat-data problems, as both the non-Bayesian formulation of XOLON and the Bayesian alternative of PFCSVM fail to achieve the classification accuracies of RFVM.

In terms of FS, although in average SVM-$\ell$1 yields a slightly better result, RFVM is able to find a more compact set of  relevant features in more datasets. Finally, in terms of sparsity in the dual variables, the results achieved by RFVM are significantly better than those of the other methods. 

From the clinical point of view, it is worth remarking that the sparsity in terms of dual variables achieved by SVM and RFVM are qualitatively different. The subjects selected by SVM are SVs, that is, observations close to the classification boundary, while the RVs selected by RFVM can live anywhere in the input space. This way, RFVM offers an alternative view to refine the population of the clinical study with a more compact and less constrained way of assessing the relevance of the subjects for the definition of the model.

\subsection{Detection of relevant features}
\label{sec:medical}

In medical applications, the detection of the variables that are relevant for the diagnosis is crucial to both describe and characterize the disease and to reduce the number of medical tests in further studies which entails high costs and inconvenience for patients. The empirical results show that RFVM is capable of capturing a reduced set of relevant variables, which  amounts to approximately  a 1$\%$ of the original set while keeping a competitive classification accuracy. To illustrate the practical usefulness of this selection, this subsection discusses in more detail the quality of the features selected for  database ALLAML. The choice of this database is motivated by two primary considerations: (1) its widespread use in leukemia research and (2)  the existence of comprehensive documentation publicly available.

The ALLAML dataset is designed for the classification of two different types of leukemia: acute lymphoblastic leukemia (ALL) and acute myeloid leukemia (AML). It comprises 7129 features representing genetic expression values extracted from microarray studies and encompasses 72 samples. These 7129 genes were selected using expert knowledge of oncologists and published in 1999 \cite{golub1999molecular}.

The use of 5-fold cross validation to obtain the experimental results, added to the fat-data characteristic of the dataset, motivates that there is not a unique set of relevant features. In fact, cross-dependencies and redundancies among features bring as a consequence that similar accuracies can be obtained by models that exploit different sets of features. To overcome this situation, the discussion is based on the examination of all the features found relevant in any of the 5 folds. 

 \begin{table}[ht!]
     \centering
     \begin{tabular}{c c c c }
     \hline
          Gene description &  Weight & Cite \\
          \hline
          \textbf{Zyxin} & +7723.06    &  \cite{golub1999molecular}\\
          \textbf{DF D component of complement} & +6315.26    & \cite{golub1999molecular}  \\
          \textbf{MPO Myeloperoxidase}& +5600.81    &  \cite{golub1999molecular}\\
          \textbf{CD33 antigen} & +5535.30    & \cite{golub1999molecular} \\
          GPX1 Glutathione peroxidase 1& +3534.11   & \cite{xiao2020lncrna} \\
          Nucleoside-diphosphate kinase& +3531.02    &  \cite{okabe1992identity}\\
          \textbf{LEPR Leptin receptor}& +3051.84   & \cite{golub1999molecular} \\
          \textbf{Epb72 gene exon 1}& +1247.82   & \cite{golub1999molecular} \\
          \textbf{Azurocidin gene}& +917.58   & \cite{golub1999molecular} \\
          Cell division control related protein & +520.65   & \cite{schnerch2012cell}  \\
          \hline
          MPL gene & -46.82    & \cite{sasaki1995expression}  \\
          GYPE Glycophorin E& -50.27    &\cite{rearden1985glycophorin} \\
          CD19 antigen& -52.87    & \cite{tasian2015cd19} \\
          \textbf{Oncoprotein 18}& -400.10    & \cite{golub1999molecular} \\
          GOT1 Glutamic-oxaloacetic transaminase 1& -613.38    & \cite{cheng2020upregulation}\\
          YWHAH & -911.46    &  \cite{namkoong2006bone}\\
          GB DEF = NADP+-dependent malic enzyme& -912.75    & \cite{chen2023targeting} \\
          MEF2A gene& -918.28    & \cite{maura2015association} \\
          BCL2 B cell lymphoma protein 2& -2003.88    & \cite{davids2012targeting} \\
          Skeletal muscle abundant protein& -2403.31    & \cite{kurek1997role} \\
          \hline
     \end{tabular}
     \caption{Relevant expression genes along all data folds. The genes are sorted from maximum to minimum weight (top to bottom). Also, the top and bottom 10 relevant features from the selected subsets are represented.
     }
     \label{tab:gen_total}
 \end{table}

Table \ref{tab:gen_total} shows the most relevant genes selected when looking the associated relevances along the 5 folds. We employed the weight vector $\mathbf{w}$, which is derived from the product $\vlam\Xst\as$, to arrange the features in descending order of relevance. The inclusion of both positive and negative values within $\mathbf{w}$ serves the purpose of identifying pertinent features for each class; this way, as AML is denoted with label 1, positively weighted features are indicative of AML occurrence, while negatively weighted features correspond to ALL\footnote{It is possible due to the type of data that compose ALLAML dataset: the data  takes positive values and the labels are binary. Hence, the signs of the weights are directly related with the classes.}. Since the paper is not intended to be a clinical study, we limit the analysis to corroborate our ranking with the literature. Besides, the third column of the table offers references to medical studies that prove the association between the selected genes and the occurrence of the leukemic disease. Furthermore, the genes highlighted in bold are also included in the study that generated the database \cite{golub1999molecular} as the top 50 most relevant genes to distinguish between ALL and AML. 

Also, in order to demonstrate the sign assignments and the relevance of the results obtained, it is worthy to point out some observations beyond the references posted at the table:
\begin{enumerate}
    \item Some selected genes (bold faced) belong to the top 50 most relevant genes to diagnose between both diseases by the study researchers \cite{golub1999molecular}. So, the model outcomes seem to match a huge part of the expert knowledge.

    \item Glutathione peroxidase 1 is involved in both AML and ALL events \cite{wei2020identification}. Upregulation is associated with the development of AML, while downregulation is linked to ALL. Thus, assigning a positive weight to this gene highlights its importance in terms of abundance for AML and its absence for ALL \cite{wei2020identification}.

    \item The gene associated with cell division appears to play a crucial role in distinguishing between AML and ALL \cite{schnerch2012cell}. It has been proven that AML shows a notably more aggressive cell proliferation, which justifies the fact that the model assigns a positive weight to this gene.

    \item The upregulation of Glycophorines (GYP) A and B is shown to be strongly related to the prevalence of ALL \cite{rearden1985glycophorin}. However, our model highlighted the relevance of GYP E instead of A and B. Nevertheless, some investigations carried out by the National Institutes of Health (NIH) reveal the close relationship between these three proteins \cite{national} and some studies connect them to Guanylate-binding proteins (GPB) through the mutation and duplication of an ancestral gene, a group of proteins also relevant in the diagnosis of ALL \cite{luo2021guanylate}.

    \item The human mitochondrial NAPD$^{+}$ dependent malice enzime, which is involved in the regulation of the NAD$^{+}$/NADH (redox balance), it was selected as relevant with a negative weight. This weight appears to be associated with the occurrence of AML rather than ALL, as its downregulation is connected to the onset of AML \cite{chen2023targeting}. 

    \item The skeletal muscle abundant protein exhibits the most negative weight. Nevertheless, we encountered some difficulties when searching the exact identity of this protein since it does not correspond to any specific skeletal protein. Although there are evidences about the harmful effects of some inhibitory factors produced during leukemic metastatic processes on skeletal muscle regeneration \cite{luo2021guanylate}, the lack of information hinders a comprehensive analysis of its biological relevance in this context.
\end{enumerate}

\subsection{Computational cost analysis}
\label{sec:Comp_cost}

The last part of this section is devoted to the analysis of the computational cost of RFVM in different fat-data scenarios. For this purpose, RFVM is evaluated in a synthetic dataset with $N=300$ and $D$ taking values in the range $[50,13500]$ (explored in logarithmic scale). Just $10\%$ of these $D$ features are informative for the classification task. Ten independent simulations were run for each value of $D$ in order to get statistically sound results.

All the simulations were run individually on the same machine with the next specifications: 2 Intel Xeon 64 E5-2640v4 CPU with 10 cores per CPU,  128Gb of RAM memory, a Linux Gentoo-compiled distribution and a capacity of 725.0649 GFlops (Test Intel Linpack).

\begin{figure}[H]
    \centering
    \includegraphics[width=1\linewidth]{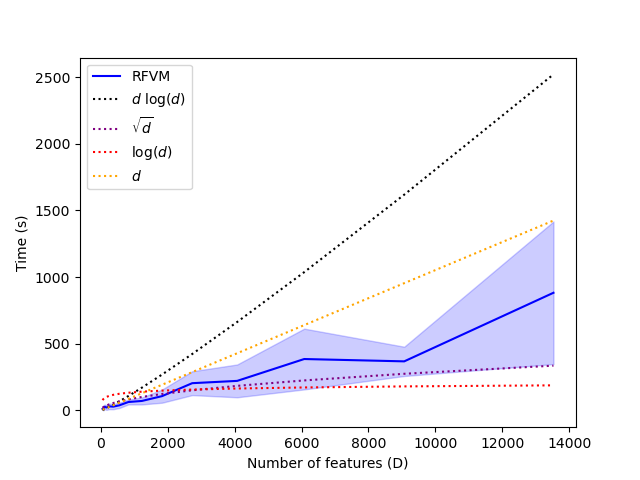}
    \caption{Computational cost of RFVM as a function of the number of input features. The dotted lines represent different complexities and the blue continuous line represents the average computational time of the proposed model. The shaded region surrounding the curve represents its standard deviation.}
    \label{fig:CC_feat}
\end{figure}

Figure \ref{fig:CC_feat} shows that RFVM exhibits a sublinear complexity between $O(d)$ and $O(\sqrt{d}$), with the confidence intervals falling between these two complexities. This desirable characteristic prevents us from facing with unfeasible execution times when dealing with high dimensional datasets. Therefore, the model does not impose time constraints beyond the inherent dimensional limitations of the dataset.

\section{Conclusions}
\label{sec:Conclusions}

This paper  has presented the Relevance Feature Vector Machine (RFVM), a novel ML model designed to  help with the management of medical prospective studies within fat-data scenarios. This model includes some new capabilities to address challenges associated with the fat-data problem such as the two-way sparsity to eliminate irrelevant or redundant information, the combined representation of samples and features over the primal and dual space, and a Bayesian framework formulation to exploit the Bayesian averaging modeling advantages when working with a reduced cohort of samples, mitigating overfitting. 

The experimental results illustrate the benefits of the model when working within fat-data scenarios. That is, in comparison with the rest of baselines, the RFVM excels at obtaining the most compact data representation while maintaining a competitive classification performance. That is, even those models that yield equal performance scores (such as the RF and the SVM-$\ell$1), they struggle to efficiently handle this type of data. Notably, the results from the ALLAML experiment indicate that the feature selection process is meaningful, as it identifies a subset of genes relevant to the development of the disease. Additionally, the model exhibits a sublinear computational burden as the number of features increases, offering significant computational efficiency when working within fat-data scenarios.

These findings highlight the utility of the RFVM in helping manage fat-data medical prospective studies. The model effectively summarizes data into a compact subset of features and samples, addressing the classification problem while eliminating the necessity of carrying out irrelevant medical tests. Moreover, it provides a descriptive patient cohort distribution, what can help manage the patient recruitment procedure. 
\\

\noindent\textbf{CRediT authorship contribution statement}
\\

\textbf{Albert Belenguer-Llorens: }Conceptaulization, Data curation, Methodology, Investigation, Validation, Software,Writing - original draft. \textbf{Carlos Sevilla-Salcedo: }Conceptualization, Methodology, Investigation, Software,Writing - review $\&$ editing \textbf{Emilio Parrado-Hernandez: }Conceptualization, Funding acquisition, Software, Supervision, Writing - review $\&$ editing \textbf{Vanessa Gomez-Verdejo: }Conceptualization, Investigation, Funding acquisition, Project administration, Supervision, Writing - review $\&$ editing. All authors have read and agreed to the published version of the manuscript.
\\

\noindent\textbf{Declaration of competing interest }
\\

The authors declare that they have no known competing financial interests or personal relationships that could have appeared to influence the work reported in this paper. 
\\

\noindent\textbf{Data availability }
\\

The authors do not have permission to share data. 
\\

\noindent\textbf{Acknowledgements}
\\

This work is partially supported by grants PID2020-115363RB-I00 funded by MCIN/AEI/10.13039/50110001103, Spain. V.G-V and C.S-S's work is partially supported by TED2021-132366B-I00 funded by MCIN/AEI/10.13039/501100011033 and by the "European Union NextGenerationEU/PRTR".

\newpage

\appendix
\setcounter{equation}{0}
\setcounter{figure}{0}
\setcounter{table}{0}
\setcounter{section}{0}
\makeatletter
\renewcommand{\theequation}{A.\arabic{equation}}
\renewcommand{\thefigure}{A\arabic{figure}}

\section{Folded Normal distribution}
\label{sec_ApFolded}

The Normal Folded distribution \cite{tsagris2014folded}, $\N_F$, defines its pdf as
\begin{equation}
    p(x|\mu,\sigma^2) = \frac{1}{\sqrt{2\pi\sigma^2}}e^{-\frac{(x-\mu)^2}{2\sigma^2}} + \frac{1}{\sqrt{2\pi\sigma^2}}e^{-\frac{(x+\mu)^2}{2\sigma^2}}, x \geq 0,
\end{equation}
where $\mu \in \R$ and $\sigma^2 > 0$ (as the univariate\footnote{Note that we defined the univariate normal distribution as we will not use the multivariate in our model inference} Normal distribution). Note that if $\mu = 0$ it is a half-normal distribution. The mean $\tilde{\mu}$ and variance $\tilde{\sigma}^2$ can be expressed as:
\begin{equation}
    \tilde{\mu} = \sqrt{\frac{2}{\sigma^2\pi}}e^{-\frac{\mu}{2\sigma^2}} + \mu \left(1 - 2\phi\left(-\frac{\mu}{\sigma}\right) \right)
\end{equation}
and
\begin{equation}
    \tilde{\sigma}^2 = \mu^2 + \sigma^2 - \tilde{\mu}^2,
\end{equation}
being $\phi$ the normal cumulative distribution function. Also, the cumulative distribution (CDF) can be expressed as:
\begin{equation}
    F(x|\mu, \sigma^2) = \frac{1}{2}\left[erf\left(\frac{x + \mu}{\sqrt{2\sigma^2}}\right) + erf\left(\frac{x - \mu}{\sqrt{2\sigma^2}}\right)\right],
\end{equation}
being $erf$ the error function:
\begin{equation}
    erf(x) = \frac{2}{\sqrt{\pi}}\int_{0}^{x} e^{-t^2}dt.
\end{equation}

Finally, the log-likelihood of the distribution can be expressed as:
\begin{equation}
    l = -\frac{N}{2}\ln(2\pi\sigma^2) - \sum^N_{n}\frac{(x_n - \mu)^2}{2\sigma^2} + \sum^N_{n}\ln\left(1 + e^{-\frac{2\mu x_n}{2\sigma^2}}\right).
\end{equation}

\appendix
\setcounter{equation}{0}
\setcounter{figure}{0}
\setcounter{table}{0}
\setcounter{section}{1}
\makeatletter
\renewcommand{\theequation}{B.\arabic{equation}}
\renewcommand{\thefigure}{B\arabic{figure}}

\section{Logistic Regression Bound}
\label{sec_logbound}

As we work within a binary framework, the probability of the logistic regression $p(t_n|\yn) = e^{\yn t_n}\sigma(-\yn)$ can be expressed as
\begin{equation}
    p(t_n = 1|\yn) = e^{\yn}\sigma(-\yn) = 
    (1+e^{-\yn})^{-1}.
\end{equation}

Also, we can express the log-likelihood of $p(t_n = 1|\yn)$ as
\begin{equation}
\label{eq:E2}
    \ln p(t_n = 1|\yn) = - \ln(1+e^{-\yn}) = \frac{\yn}{2} - \ln(e^{\yn}+e^{-\yn}),
\end{equation}
where $f(\yn)=-\ln(e^{\yn}+e^{-\yn})$ is a convex function in $\yn^2$. As demonstrated at \cite{jaakkola2000bayesian}, we can use a curve to lower bound a convex function. Hence, we lower bound $f(\yn)$ with the first-order Taylor series expansion in $\yn^2$ as
\begin{equation}
\label{eq:E3}
    f(\yn) \geq f(\xi) + \frac{\partial f(\xi_n)}{\partial \xi_n^2}(\yn^2 - \xi_n^2) = -\frac{\xi_n}{2} + \ln p(t_n = 1|\xi_n) + \frac{1}{4\xi_n} tgh\left(\frac{\xi}{2}\right)(\yn^2 - \xi_n^2),
\end{equation}
where $\xi_n^2$ is the center of the Taylor serie and $f(\yn) = f(\xi_n) + \frac{\partial f(\xi_n)}{\partial \xi_n^2}(\yn^2 - \xi_n^2)$ when $\xi_n^2 = \yn^2$. Now, as presented in \cite{jaakkola2000bayesian}, if we combine Eq. (\ref{eq:E2}) and Eq. (\ref{eq:E3}) we achieve the lower bound over $\ln p(t_n = 1|\yn)$:
\begin{equation}
    \ln p(t_n = 1|\yn) = 
    e^{\yn}\sigma(-\yn) \geqslant h(\yn, \xi_n) = 
    e^{\yn t_n}\sigma(\xi_n)e^{-\frac{\yn + \xi_n}{2} - \lambda(\xi_n)(\yn^2 - \xi_n^2)},
\end{equation}
where $\lambda(a) = \frac{1}{2a}(\sigma(a) - \unmed)$.

\appendix
\setcounter{equation}{0}
\setcounter{figure}{0}
\setcounter{table}{0}
\setcounter{section}{2}
\makeatletter
\renewcommand{\theequation}{C.\arabic{equation}}
\renewcommand{\thefigure}{C\arabic{figure}}

\section{Variational Inference}
\label{sec_Ap1}
This section presents the mathematical developments  of the approximated posterior distributions of the model variables, i.e., $q(\mathbf{\Theta})$. As presented in Figure \ref{fig:GM_RFVM_class}, the model variables follow these prior distributions $p(\mathbf{\Theta})$:
\begin{equation}
    p(t_n|\yn) = e^{\yn t_n}\sigma(-\yn)
\end{equation}
\begin{equation}
    \yn \sim \N(\xn\vlam\Xst\as + b , \tau^{-1})
\end{equation}
\begin{equation}
    \an \sim \N(0, \fin^{-1})
\end{equation}
\begin{equation}
    \fin \sim \G(\anfi, \bnfi)
\end{equation}
\begin{equation}
    \vd \sim \N_{F}(0,\alfad^{-1})
    \label{eq:folded}
\end{equation}
\begin{equation}
    \alfad \sim \G(\adalfa,\bdalfa)
\end{equation}
\begin{equation}
    b \sim \N(0,1)
\end{equation}
\begin{equation}
    \tau \sim \G(\atau, \btau)
\end{equation}

Hence, following the equations in subsection \ref{sec:inference}, we can determine the approximate posterior of all these variables by means of mean-field variational inference. That is, maximizing the $L(q)$ presented in Eq. \eqref{eq:ELBO} for each loop iteration and applying the analytical solution of each variable extracted from Eq. \eqref{eq:lnqopt}. 

To clarify the relationships between the model variables $\mathbf{\Theta}$, and apply Eq. \eqref{eq:lnqopt}, we can define the joint model distribution as:
\begin{equation}
    p(\mathbf{t},\mathbf{X},\boldsymbol{\Theta}) = p(\mathbf{t}|\ys)p(\ys|\X, \Xs, \vs, \as, \taus, b)p(\as|\fis)p(\fis)p(\vs|\alfas)p(\alfas)p(\taus)p(b).
\end{equation}

Furthermore, as some variables need to calculate the expectation over the output distribution $p(\ys|\X, \Xs, \vs, \as, \taus, b)$, let us approximate the log-probability for subsequent developments:
\begin{equation}
\begin{split}
    & \ln p(\ys|\X, \Xs, \vs, \as, \taus, b)  \thickapprox  \sum_{n=1}^{N}\ln \N(\xn\vlam\Xst\as + b, \taus^{-1}) \\
    & = \sum_{n=1}^{N}(-\unmed\lnpi + \unmed\lntau -\frac{\taus}{2}(\yn - \xn\vlam\Xst\as -b)^\top(\yn - \xn\vlam\Xst\as - b))\\
    &= \sumn(-\unmed\lnpi + \unmed\lntau - \frac{\taus}{2}(\yn^2 - \yn\xn\vlam\Xst\as - \yn b -(\xn\vlam\Xst\as)^\top\yn\\ & + \as^T\Xs\vlam\xnt\xn\vlam\Xst\as +\as^T\Xs\vlam\xnt b - b\yn + b\xn\vlam\Xst\as +b^2))\\
    & = \sumn(-\unmed\lnpi + \unmed\lntau - \frac{\taus}{2}(\yn^2 - 2\yn\xn\vlam\Xst\as-2\yn b\\ &+ 2\mathbf{a}^\top\Xs\vlam\xnt b + b^2 +\as^\top\Xs\vlam\xnt\xn\vlam\Xst\as)).
    \label{eq:aa}
\end{split}
\end{equation}

\begin{flushleft}
\textbf{Distribution of a}
\end{flushleft}

Following the mean-field procedure defined in Eq. \eqref{eq:lnqopt}, we can define the approximated posterior distribution of $\mathbf{a}$ as:
\begin{equation}
    \ln q_{\mathbf{a}}(\as) = \E_{\taus,\vs,b, \mathbf{y}}[\ln p(\ys|X, \Xs, \taus, \as, \vs, b)] + \E_{\fis}[\ln p(\as, \fis)] + const.
\end{equation}

Hence, we can express the first term of the summation as
\begin{equation}
\begin{split}
    & \ln p(\ys|X, \Xs, \taus, \as, \vs, b) =
    \sumn \taus[\yn\xn\vlam\Xst\as - \as^\top\Xs\vlam\xnt b - \\ & \unmed\as^\top\Xs\vlam\xnt\xn\vlam\Xst\as] + const\\
    &= \sumn \taus[\yn\vs\xlam\Xst\as - b\vs\xlam\Xst\as - \unmed\as^\top\Xs\xlam\vs\vs^\top\xlam\Xst\as] + const,
    \label{eq:a1}
\end{split}
\end{equation}
where some terms will become constant due to the following expectation. Also, we can express the second term as
\begin{equation}
\begin{split}
    \ln p(\as, \fis) & =
    \sumns(-\unmed\lnpi + \unmed\ln(\fin) - \frac{\fin}{2}\an^2) + const\\
    &= -\unmed\as^\top\filam\as + const.
    \label{eq:a2}
\end{split}
\end{equation}

Finally, if we join both Eq. \eqref{eq:a1} and \eqref{eq:a2} and apply the expectations, we get
\begin{equation}
\begin{split}
    \ln q_{\as}(\as) = & \sumn \tauE[\langle\yn\rangle\vE\xlam\Xst\as - b\vE\xlam\Xst\as - \unmed\as^\top\Xs\xlam\vvt\xlam\Xst\as] \\ &-\unmed\as^\top\filamE\as +const. 
\end{split}
\end{equation}

Comparing the expression obtained with the desired posterior distribution (normal distribution), we can identify terms to obtain the
\begin{equation}
    q_{\as}(\as) = \N(\as|\aE,\acov^{-1}),
\end{equation}
where
\begin{equation}
    \acov^{-1} = \tauE \sumn(\Xs\xlam\vvt\xlam\Xst) + \filamE
\end{equation}

\begin{equation}
    \aE = \tauE(\langle\ys\rangle^\top - \mathbf{1}_{N}\bE)X\vlamE\Xst\acov,
\end{equation}

\begin{flushleft}
\textbf{Distribution of v}
\end{flushleft}

Following the same idea presented in $\as$, we can define the approximated posterior distribution of $\vs$ as
\begin{equation}
    \ln q_{\vs}(\vs) = \E_{\taus,\as, b}[\ln p(\ys|X, \Xs, \taus, \as, \vs, b)] + \E_{\alfas}[\ln p(\vs|\alfas)] + const.
\end{equation}

As the elements of $\mathbf{v}$ are independent, the first term of the summation can be expressed as
\begin{equation}
\begin{split}
    &\ln p(\ys|X, \Xs, \taus, \as, \vs, b) = \sumn\sumd(\taus(\yn - b)\vd\xnd\xsdt\as)\\ &- \sumn\sumduno\sumddos(\frac{\taus}{2}\as^\top\tilde{\mathbf{x}}_{:,d1}x_{n,d1}v_{d1}v_{d2}x_{n,d2}\tilde{\mathbf{x}}_{:,d2}\as) + const.\\
    &= \sumn\sumd(\taus(\yn - b)\vd\xnd\xsdt\as)
    -\sumn\sumd(\frac{\taus}{2}\as^\top\xsd\xnd^2\vd^2\xsdt\as)\\
    &-\sumn\sumduno\underset{(d1 \neq d2)}{\sumddos}(\frac{\taus}{2}\as^\top\tilde{\mathbf{x}}_{:,d1}x_{n,d1}v_{d1}v_{d2}x_{n,d2}\tilde{\mathbf{x}}_{:,d2}\as) + const.
    \label{eq:a3}
\end{split}
\end{equation}

Now, we can develop the second term of the summation, $\ln p(\vs|\alfas)$. But, as we can appreciate from Eq. \eqref{eq:folded}, each $\vd$ follows a folded Normal distribution $\N_{F}$, i.e., 
\begin{equation}
    p(\vd|\mu,\sigma^2) = \frac{1}{\sqrt{2\pi\sigma^2}}e^{-\frac{(\vd-\mu)^2}{2\sigma^2}} + \frac{1}{\sqrt{2\pi\sigma^2}}e^{-\frac{(\vd+\mu)^2}{2\sigma^2}}, \vd \geqslant 0.
\end{equation}

Hence, as required in Eq. \eqref{eq:lnqopt}, we applied the logarithm obtaining
\begin{equation}
    \ln p(\vs|\mu,\sigma^2) = -\frac{N}{2}\lnpi + \frac{N}{2}\ln(\sigma^2) - \sumd \frac{(\vd - \mu)^2}{2\sigma^2} + \sumd \ln(1 + e^{\frac{-2\mu\vd}{\sigma^2}}) + const.
\end{equation}
where, as the prior of $\vd$ (Eq. \eqref{eq:folded}) defines $\mu = 0$, the last term of the summation can be expressed as $\sumd \ln(1 + e^{\frac{-2\mu\vd}{\sigma^2}}) = \sumd \ln(2)$, that is, a constant value. Thus, the distribution will be finally expressed as
\begin{equation}
    \ln p(\vs|0,\sigma^2) = -\frac{N}{2}\lnpi + \frac{N}{2}\ln(\sigma^2) - \sumd \frac{(\vd)^2}{2\sigma^2}  + const.
\end{equation}
which is a log-Normal distribution.

Therefore, once $\ln p(\vs|\alfas)$ can be treated as a log-Normal distribution, we can easily develop $\E_{\alfas}[\ln p(\vs|\alfas)]$ like in Eq. \eqref{eq:a2} as
\begin{equation}
    \ln p(\vs|\alfas) = \sumd(-\frac{\alfad}{2}\vd^2) = \sumd(\vd(-\frac{\alfad}{2})\vd) + const.
    \label{eq:a4}
\end{equation}

If we join both Eq. \eqref{eq:a3} and \eqref{eq:a4} and apply expectations, we get the posterior
\begin{equation}
    q_{\vd}(\vd) = \N_{F}(\vd|\langle\vd\rangle,\sigd^2),
\end{equation}
where
\begin{equation}
    \langle\vd\rangle = (\sigd^{-2})\sumn[\tauE(\langle\yn\rangle - \bE)\xnd\xsdt\aE - \underset{(d \neq \hat{d})}{\sumdrara}[\tauE\langle v_{\hat{d}}\rangle \tilde{\mathbf{x}}_{:,\hat{d}}^\top\aat\xsdt\xnd x_{n,\hat{d}}]]
\end{equation}

\begin{equation}
    \sigd^2 = \tauE\xsdt\aat\xsd\sumn(\xnd^2) + \langle\alfad\rangle.
\end{equation}

Also, as $q(\vd)$ is a folded Normal distribution, its parameters will be iteratively updated by the following rules:
\begin{equation}
    \hat{\vdE} = \sqrt{\frac{2}{\sigd^2\pi}}e^{-\frac{\vdE}{2\sigd^2}} + \vdE(1 - 2\phi(-\frac{\vdE}{\sigd}))
\end{equation}

\begin{equation}
    \tilde{\sigd^2} = \vdE^2 + \sigd^2 - \hat{\vdE}^2,
\end{equation}
being $\phi$ the normal cumulative distribution function.

\begin{flushleft}
\textbf{Distribution of $\bm{\delta}$}
\end{flushleft}

In this case, following the mean-field rule, the posterior approximation can be defined as
\begin{equation}
    \ln q_{\alfas}(\alfas) = \E_{\vs}[\ln p(\vs|\alfas)] + \E[\ln p(\alfas)] + const.
\end{equation}

Following previous developments, we applied the expectation over both terms. The first term can be presented as
\begin{equation}
\begin{split}
    & \E_{\vs}[\ln p(\vs|\alfas)] = \E\left[\sumd(-\unmed\lnpi + \unmed\ln(\alfad) - \frac{\alfad}{2}\vd^2)\right] \\
    & = \sumd(\unmed\ln(\alfad) - \unmed\langle v_{d}^2\rangle\alfad) = \sumd(\unmed\ln(\alfad)) - \unmed\vvt\sumd(\alfad) + const,
\end{split}
\end{equation}
while the second can be expressed as
\begin{equation}
    \E[\ln p(\alfas)] = \sumd(-\bceroalfa\alfad + (\aceroalfa-1)\ln(\alfad)) + const.
\end{equation} 

So, if we join both terms, the final distribution can be written as
\begin{equation}
    q_{\alfas}(\bm{\delta}) = \Gamma(\bm{\delta}|\bm{\alpha}^{\delta}, \bm{\beta}^{\delta}),
\end{equation}
where
\begin{equation}
    \aalfa = \unmed + \aceroalfa
\end{equation}

\begin{equation}
    \balfa = \bceroalfa + \unmed\Lambda_{\vvt}.
\end{equation}

\begin{flushleft}
\textbf{Distribution of $\psi$}
\end{flushleft}

The approximated posterior distribution can be derived from
\begin{equation}
    \ln q_{\fis}(\fis) = \E_{\as}[\ln p(\as|\fis)] + \E[\ln p(\fis)].
\end{equation}

Note that this equation can be solved following the same procedure as $\alpha$, but applying the expectation over different variables. Hence, the final distribution will be equivalent to the one obtained of $\bm{\delta}$, that is
\begin{equation}
    q_{\fis}(\bm{\psi}) = \Gamma(\bm{\psi}|\bm{\alpha}^{\psi}, \bm{\beta}^{\psi}),
\end{equation}
where
\begin{equation}
    \afi = \unmed + \acerofi
\end{equation}

\begin{equation}
    \bfi = \bcerofi + \unmed diag(\aat).
\end{equation}

\begin{flushleft}
\textbf{Distribution of $\taus$}
\end{flushleft}

We determine the posterior distribution of the noise $\mathbf{\taus}$ by solving the approximated posterior
\begin{equation}
    \ln q_{\taus}(\taus) = \E_{\vs, \as, b}[\ln p(\ys|X,\Xs,\vs,\as\taus,b)] + \E[\ln p(\taus)].
\end{equation}

Following the development of Eq. \eqref{eq:aa}, we operate the first term as
\begin{equation}
\begin{split}
 & \E_{\vs, \as, b}[\ln p(\ys|X,\Xs,\vs,\as\taus,b)] = \E_{\vs, \as, b}\left[\sumn(\ln \N(\xn\vlam\Xst\as + b , \tau^{-1}))\right] \\
 & = \E_{\vs, \as, b}[\sumn(\unmed\lntau - \frac{\taus}{2}(\yn^2 - 2\yn\xn\vlam\Xst\as-2\yn b \\ &+ 2\an^\top\Xs\vlam\xnt b + b^2 +\as^\top\Xs\vlam\xnt\xn\vlam\Xst\as))] + const \\
 & = \sumn(\unmed\lntau) - \frac{\taus}{2}(\langle\ys^\top\ys \rangle-2\langle\ys\rangle^\top \X\vlamE\Xst\aE \\
    &+ 2 \mathbf{1}_{N}\bE \X\vlamE\Xst\aE + \langle b^2\rangle + \sumn Trace(\vvt\xlam\Xst\aat\Xs\xlam)) + const,
\end{split}
\end{equation}
and the second term, following the same procedure as in $\alpha$ and $\psi$, can be written as
\begin{equation} 
    \E[\ln p(\taus)] = -\bcerotau\taus + (\acerotau -1)\lntau + const.
\end{equation}

Now, if we join both expectation elements and identify terms, we have that the new approximated posterior distribution of the noise can be expressed as
\begin{equation}
    q_{\taus}(\taus) = \Gamma(\taus|\alpha^{\taus}, \beta^{\taus}),
\end{equation}
where
\begin{equation}
    \atau = \frac{N}{2} + \acerotau
\end{equation}

\begin{equation}
\begin{split}
    & \btau = \bcerotau + \unmed(\langle\ys^\top\ys\rangle -2\langle\ys\rangle^\top \X\vlamE\Xst\aE
    + 2\mathbf{1}_{N}\bE \X\vlamE\Xst\aE + \langle b^2\rangle \\ & +\sumn Trace(\vvt\xlam\Xst\aat\Xs\xlam) -2\langle\ys\rangle^\top \mathbf{1}_{N}^\top\bE^\top).
\end{split}
\end{equation}

\begin{flushleft}
\textbf{Distribution of b}
\end{flushleft}

The bias posterior distribution can be derived from
\begin{equation}
    \ln q_{b}(b) = \E_{\vs, \as, \taus}[\ln p(\ys|X, \Xs, \vs, \as, \taus, b)] + \E[\ln p(b)] + const.
\end{equation}

Following the ideas presented along this section, we develop the first summation term as
\begin{equation}
\begin{split}
    &\E_{\vs, \as, \taus}[\ln p(\ys|X, \Xs, \vs, \as, \taus, b)] = \E_{\vs, \as, \taus}\left[\sumn(-\frac{\taus}{2}(-2\yn b + 2\as^\top\Xs\vlam\xnt b + b^2))\right] \\
    & = \tauE Trace(\Lambda_{\langle\ys\rangle})b - \tauE b\mathbf{1}_{N}X\vlam\Xst\aE - \unmed N\tauE b^2 + const,
\end{split}
\end{equation}
and the second
\begin{equation}
    \E[\ln p(b)] = -\unmed b^2 + const.
\end{equation}

If we join both expectations and identify elements, we get
\begin{equation}
    q_{b}(b) = \N(b| \langle b \rangle, \Sigma_b),
\end{equation}
with
\begin{equation}
    \Sigma_{b}^{-1} = N \tauE +1
\end{equation}

\begin{equation}
    \bE = \tauE[-F_{1}X\vlamE\Xst\aE + Trace(\Lambda_{\langle y\rangle})]\Sigma_{b}.
\end{equation}

\begin{flushleft}
\textbf{Distribution of y}
\end{flushleft}

Finally, we have to infer the distribution of $\mathbf{y}$. To do so, we follow \cite{bishop2006pattern} in order to be able to apply a mean-field inference procedure framework over $\mathbf{y}$, we define a lower bound by applying the second-order Taylor-series expansion over $p(t_n|\yn)$, having
\begin{equation}
    p(t_n|\yn) = 
    e^{\yn t_n}\sigma(-\yn) \geqslant h(\yn, \xi_n) = 
    e^{\yn t_n}\sigma(\xi_n)e^{-\frac{\yn + \xi_n}{2} - \lambda(\xi_n)(\yn^2 - \xi_n^2)},
\end{equation}
which allows us to merge it within Eq. \eqref{eq:lnqopt} to drive to a suboptimal parameterization of a normal distribution of the approximated posterior $q(\mathbf{y})$.

Hence, following the mean field approximation, we can write $q(\mathbf{y})$ as
\begin{equation}
    \ln q_{\ys}(\ys) = \E[\ln p(\ys |X, \Xs, \as, \vs, b, \taus)] + \E[\ln h(\ys, \mathbf{\xi})].
\end{equation}

As in previous inferences, we can treat both terms independently. In that case, the first term can be easily inferred following the same procedure as in Eq. \eqref{eq:a1}. However, the second term needs to be simplified while applying the expectation over it as
\begin{equation}
\begin{split}
    \E[\ln (h(\yn, \xi_n))] & = \E\left[\sumn(\ln(\sigma(\xi_n)) + \yn t_n - \unmed(\yn + \xi_n) - \lambda(\xi_n)(\yn^2 - \xi_n^2))\right] \\
    & = \sumn(\yn t_n - \unmed \yn - \lambda(\xi_n)\yn^2) + const \\
    & = \sumn((t_n - \unmed)\yn - \yn^2 \xi_n) + const.
\end{split}
\end{equation}

Hence, we join both terms to achieve
\begin{equation}
\begin{split}
    q(\yn)  & = \E\left[\sumn[(t_n - \unmed)\yn - \xi_n\yn^2 - \taus\unmed\yn^2 + \taus\xn\vlam\Xst\as\yn + \taus b\yn]\right] \\
    & = \sumn\left[(t_n - \unmed + \tauE\xn\vlamE\Xst\aE + \tauE\bE)\yn - (\xi_n + \tauE\unmed)\yn^2\right].
\end{split}
\end{equation}

This way, we can obtain the approximated posterior output distribution as
\begin{equation}
    q_{\ys}(\mathbf{y}) = \N(\mathbf{y}|\langle \mathbf{y} \rangle, \Sigma_{\mathbf{y}}),
\end{equation}
where
\begin{equation}
    \Sigma_{\ys}^{-1} = \tauE I +2\Lambda_{\xi}
\end{equation}

\begin{equation}
    \langle \ys\rangle = (\mathbf{t}\mathbf{1}_{N} - \unmed\mathbf{1}_{N} + \tauE\aE\Xs\vlamE X^\top + \tauE\bE \mathbf{1}_{N}) \Sigma_{\ys}.
\end{equation}

Also, to calculate the variational parameter $\bm{\xi}$, we have to maximize the ELBO. In this case, we only have to take into account the term that include $\bm{\xi}$ ($\ln(h(\mathbf{y}, \bm{\xi}))$), defined as
\begin{equation}
    \E_q(\ln(h(\mathbf{y}, \bm{\xi}))) = \sumn(\ln(\sigma(\xi_n)) + \langle \yn\rangle t_n - \unmed(\langle \yn\rangle + \xi_n) - \lambda(\xi_n)(\E[\yn^2] - \xi_n^2)).
\end{equation}

Hence, in order to find the maximum value, we can derivate $\E_q(\ln(h(\mathbf{y}, \bm{\xi})))$ with respect to each $\xi_n$ and equal to zero:
\begin{equation}
    \frac{\partial\E_q(\ln(h(\mathbf{y}, \bm{\xi})))}{\partial\xi_n} = \lambda^{'}(\xi_n)(\E[\yn^2]-\xi_n^2) = 0,
\end{equation}
where the derivative function $\lambda^{'}(\xi_n)$ is monotonic for $\xi_n   \geqslant 0$, and we can focus on non-negative values of $\bm{\xi}$ due to its symmetry around 0:
\begin{equation}
    \lambda^{'}(\xi_n) \neq 0 \longrightarrow \xi_n^{new2} = \E[\yn^2] = \langle \yn^2 \rangle +\Sigma_{\yn}.
\end{equation}

\appendix
\setcounter{equation}{0}
\setcounter{figure}{0}
\setcounter{table}{0}
\setcounter{section}{3}
\makeatletter
\renewcommand{\theequation}{D.\arabic{equation}}
\renewcommand{\thefigure}{D\arabic{figure}}

\section{Update of the lower bound $L(q)$}
\label{sec_Ap2}

In order to calculate the value of the $L(q)$ at each iteration and subsequently drive the training procedure, we can derivate the $L(q)$ expression as presented in Section \ref{sec:inference} as:
\begin{equation}
\begin{split}
    L(q) & = -\int q(\bm{\Theta}) \ln\left( \frac{q(\bm{\Theta})}{p(\bm{\Theta},\mathbf{t},\X)}\right) d\bm{\Theta} = \int q(\bm{\Theta}) \ln\left(q(\bm{\Theta})\right) 
    - \int q(\bm{\Theta})\ln\left(p(\bm{\Theta}, \mathbf{t}, \X)\right) \\
    & = \E_{q}\left[\ln (q(\bm{\Theta}))\right] - \E_{q}\left[\ln (p(\bm{\Theta}, \mathbf{t}, \X))\right].
\end{split}
\end{equation}

We will calculate separately the terms related to the expectation of the joint probability $\E_{q}\left[\ln (p(\bm{\Theta}, \mathbf{t}, \X))\right]$ and those related with the expectation of the entropy $\E_q\left[\ln(q(\mathbf{\Theta}))\right]$.
\\

\textbf{Terms associated with $\E_q\left[\ln(p(\mathbf{\Theta}, \mathbf{t}, \X))\right]$}
\\

This term comprises the following:
\begin{equation}
\begin{split}
    \E_{q}\left[\ln (p(\bm{\Theta}, \mathbf{t}, \X))\right] & = \E_q\left[\ln(p(\ys|\X, \Xs, \vs, \as, \taus, b))\right] + \E_q\left[\ln(p(\as,\fis))\right] +\E_q\left[\ln(p(\fis))\right] \\ &+ \E_q\left[\ln(p(\vs|\alfas))\right] +\E_q\left[\ln(p(\alfas))\right] +\E_q\left[\ln(p(\taus))\right] +\E_q\left[\ln(p(b))\right] \\ & + \E_q\left[\ln(p(\mathbf{t}|\mathbf{y}))\right]
\end{split}
\end{equation}

We can analyze each term independently to later merge them into the same expression.

First, the expectation over $\ln (p(\bm{\Theta}, \mathbf{t}, \X))$ can be developed as

\begin{equation}
\begin{split}
    & \E_q\left[\ln(p(\ys|\X, \Xs, \vs, \as, \taus, b))\right]  =  \E_q\left[\sum_{n=1}^{N}\ln \N(\xn\vlam\Xst\as + b, \taus^{-1})\right] \\
    & = \E_q[\sumn(-\unmed\lnpi + \unmed\lntau - \frac{\taus}{2}(\yn^2 - 2\yn\xn\vlam\Xst\as-2\yn b\\ &+ 2\an^\top\Xs\vlam\xnt b + b^2 +\as^\top\Xs\vlam\xnt\xn\vlam\Xst\as))] \\
    & = -\frac{N}{2}\lnpi + \unmed\sumn\E\left[\ln(\taus)\right] - \unmed\E[\taus]\sumn[\langle \yn^2 \rangle - 2\langle\an\rangle^\top\Xs\vlamE\xnt(\langle \yn\rangle - \langle b\rangle) \\ & + 2\langle \yn\rangle \langle b\rangle + \langle b^2\rangle + Trace(\vvt\Lambda_{\mathbf{x_{n,:}}}\Xst\aat\Xs\Lambda_{\mathbf{x_{n,:}}})].
    \label{eq:as}
\end{split}
\end{equation}

Also, we can calculate the expectation of a variable that follows a gamma distribution, as the noise, such as
\begin{equation}
    \E_q\left[\ln(\taus)\right] = \omega(\alpha^{\taus}) - \ln(\beta^{\taus}),
\end{equation}
and 
\begin{equation}
    \E_q[\taus] = \frac{\alpha^{\taus}}{\beta^{\taus}},
\end{equation}
where $\omega$ represents the digamma function, which denotes the derivative of the logarithm of the gamma function as
\begin{equation}
    \omega(\taus) = \frac{\partial\ln\Gamma(\taus)}{\partial\taus} = \frac{\Gamma'(\taus)}{\Gamma(\taus)}.
\end{equation}

Therefore, Eq. \eqref{eq:as} can be rewritten as
\begin{equation}
\begin{split}
    & \E_q\left[\ln(p(\ys|\X, \Xs, \vs, \as, \taus, b))\right] = -\frac{N}{2}\lnpi + \frac{N}{2}(\omega(\alpha^{\taus}) - \ln(\beta^{\taus}))  - \unmed\frac{\alpha^{\taus}}{\beta^{\taus}}\sumn(\langle \yn^2\rangle \\ & - 2\langle\as\rangle^\top\Xs\vlamE\xnt(\langle \yn\rangle - \langle b\rangle) + 2\langle \yn\rangle \langle b\rangle + \langle b^2\rangle + Trace(\vvt\Lambda_{\mathbf{x_{n,:}}}\Xst\aat\Xs\Lambda_{\mathbf{x_{n,:}}})).
\end{split}
\end{equation}

Also, if we look at the $\beta^{\taus}$ expression depicted in Table \ref{tab:my_label}, we can simplify the last term of the summation to achieve:
\begin{equation}
    \E_q\left[\ln(p(\ys|\X, \Xs, \vs, \as, \taus, b))\right] = -\frac{N}{2}\lnpi + \frac{N}{2}(\omega(\alpha^{\taus}) - \ln(\beta^{\taus}))  - \unmed\frac{\alpha^{\taus}}{\beta^{\taus}}(\beta^{\taus} - \beta_0^{\taus}).
\end{equation}

Also, we apply the expectation over $p(\mathbf{t}|\mathbf{y})$ as
\begin{equation}
\begin{split}
    & \E_q\left[\ln p(\mathbf{t}|\mathbf{y})\right] = \E_q[\sumn(\ln(\sigma(\xi_n)) + \yn t_n - \unmed(\yn + \xi_n) - \lambda(\xi_n)(\yn^2 - \xi_n^2))] \\ & = \sumn(\ln(\sigma(\xi_n)) + \langle \yn\rangle t_n - \unmed(\langle \yn\rangle + \xi_n) - \lambda(\xi_n)(\langle \yn^2\rangle - \xi_n^2)).
\end{split}
\end{equation}

Now, if we go forward to the main model variables, we can derive the term associated to $\as$ as

\begin{equation}
\begin{split}
    &\E_q\left[\ln(p(\as|\fis))\right] = \sum_{\tilde{n}}^{\tilde{N}}\E\left[-\unmed\lnpi + \unmed\ln(\psi_{\tilde{n}}) - \frac{\psi_{\tilde{n}}}{2}\an^2\right] \\
    & = -\frac{\tilde{N}}{2}\lnpi + \unmed\sum_{\tilde{n}}^{\tilde{N}}\E\left[\ln(\psi_n)\right] - \unmed\sum_{\tilde{n}}^{\tilde{N}}\E\left[\psi_{\tilde{n}}\right]\E\left[a_{\tilde{n}}^2\right] \\
    &=-\frac{\tilde{N}}{2}\lnpi + \unmed\sum_{\tilde{n}}^{\tilde{N}}(\omega(\alpha_{\tilde{n}}^{\psi}) - \ln(\beta_{\tilde{n}}^{\psi})) -\unmed\sum_{\tilde{n}}^{\tilde{N}}\left(\frac{\alpha_{\tilde{n}}^{\psi}}{\beta_{\tilde{n}}^{\psi}}\right)(diag(\aat))\\
    & = -\frac{\tilde{N}}{2}\lnpi + \unmed\sum_{\tilde{n}}^{\tilde{N}}(\omega(\alpha_{\tilde{n}}^{\psi})) - \unmed\sum_{\tilde{n}}^{\tilde{N}}\left(\frac{\alpha_{\tilde{n}}^{\psi}}{\beta_{\tilde{n}}^{\psi}}\left(2\left(\beta_{\tilde{n}}^{\psi} - \beta_0^{\psi}\right)\right)\right) \\
    & = -\frac{\tilde{N}}{2}\lnpi + \unmed\sum_{\tilde{n}}^{\tilde{N}}(\omega(\alpha_{\tilde{n}}^{\psi})) - \sum_{\tilde{n}}^{\tilde{N}}(\alpha_{\tilde{n}}^{\psi}) + \beta_0^{\psi}\sum_{\tilde{n}}^{\tilde{N}}\left(\frac{\alpha_{\tilde{n}}^{\psi}}{\beta_{\tilde{n}}^{\psi}}\right),
\end{split}
\end{equation}

and, following the same procedure, the term associated to $\vs$ as
\begin{equation}
    \E_q\left[\ln(p(\vs|\alfas))\right] = -\frac{D}{2}\lnpi + \unmed\sum_{d}^{D}(\omega(\alpha_{d}^{\delta})) - \sum_{d}^{D}(\alpha_{d}^{\delta}) + \beta_0^{\delta}\sum_{d}^{D}\left(\frac{\alpha_{d}^{\delta}}{\beta_{d}^{\delta}}\right) + D\ln(2),
\end{equation}
note that we added the term $D\ln(2)$ as $\mathbf{v}$ follows a folded Normal distribution with a final term $\sumd \ln(1 - e^{\frac{-2\mu\vd}{\sigma^2}})$ where $\mu = 0$.

We can also develop the gamma variables that include the sparse priors and the noise. We start with $\fis$ as
\begin{equation}
\begin{split}
    & \E_q\left[\ln(p(\fis))\right] = \sum_{\tilde{n}}^{\tilde{N}}\left(-\beta_{0}^{\psi}\E\left[\psi_{\tilde{n}}\right] + \alpha_0^{\psi}\ln(\beta_0^{\psi}) + (\alpha_0^{\psi} - 1)\E\left[\ln(\psi_{\tilde{n}})\right] - \ln\left(\Gamma(\alpha_0^{\psi})\right)\right) \\
    & = \tilde{N}(\alpha_0^{\psi}\ln(\beta_0^{\psi}) - \ln(\Gamma(\alpha_0^{\psi}))) + \sum_{\tilde{n}}^{\tilde{N}}\left(-\beta_0^{\psi}\frac{\alpha_{\tilde{n}}^{\psi}}{\beta_{\tilde{n}}^{\psi}} + (\alpha_0^{\psi}-1)(\omega(\alpha_{\tilde{n}}^{\psi}) - \ln(\beta_{\tilde{n}}^{\psi}))\right),
    \label{eq:afi}
\end{split} 
\end{equation}
to subsequently, following the same procedure, obtain $\alfas$ and the noise:
\begin{equation}
    \E_q\left[\ln(p(\alpha))\right] = D(\alpha_0^{\delta}\ln(\beta_0^{\delta}) - \ln(\Gamma(\alpha_0^{\delta}))) + \sum_{d}^{D}\left(-\beta_0^{\delta}\frac{\alpha_{d}^{\delta}}{\beta_{d}^{\delta}} + (\alpha_0^{\delta}-1)(\omega(\alpha_{d}^{\delta}) - \ln(\beta_{d}^{\delta}))\right)
\end{equation}

\begin{equation}
    \E_q\left[\ln(p(\taus))\right] = \alpha_0^{\taus}\ln(\beta_0^{\taus}) - \ln(\Gamma(\alpha_0^{\taus})) -\beta_0^{\taus}\frac{\alpha^{\taus}}{\beta^{\taus}} + (\alpha_0^{\taus}-1)(\omega(\alpha^{\taus}) - \ln(\beta^{\taus})).
\end{equation}

Finally, as the prior of the bias follows a standard normal distribution, it can be expressed as
\begin{equation}
    \E[\ln p(b)] = \E\left[-\unmed\ln(2\pi) + \unmed\ln(1) - \unmed b^2
    \right] = -\unmed\ln(2\pi) - \unmed\langle b^2\rangle.
\end{equation}
\\

\textbf{Terms associated with the entropy, $\E_q\left[\ln(q(\mathbf{\Theta}))\right]$}
\\

The entropy can be decomposed by the summation of the following terms:
\begin{equation}
\begin{split}
    \E_q\left[\ln(q(\mathbf{\Theta}))\right] & = \E_q\left[\ln q_{\as}(\mathbf{a})\right] + \E_q\left[\ln q_{\vs}(\mathbf{v})\right] + \E_q\left[\ln q_{\alfas}(\bm{\delta})\right] + \E_q\left[\ln q_{\fis}(\bm{\psi})\right] +\E_q\left[\ln q_{\taus}(\taus)\right] \\ &  + \E_q\left[\ln q_{b}(b)\right] + \E_q\left[\ln q_{\ys}(\mathbf{y})\right]
\end{split}
\end{equation}

Also, as in the $\E_q\left[\ln(p(\X, \mathbf{\Theta}))\right]$ case, the different terms can be independently analyzed as:

\begin{equation}
    \E_q\left[\ln q_{\as}(\mathbf{a})\right] = \sum_{\tilde{n}}^{\tilde{N}}\left(\unmed\ln(2\pi e) + \unmed\ln|\Sigma_{\mathbf{a}}|\right) = \frac{\tilde{N}}{2}\ln(2\pi e) + \frac{\tilde{N}}{2}\ln|\Sigma_{\mathbf{a}}|
\end{equation}

\begin{equation}
    \E_q\left[\ln q_{\vs}(\mathbf{v})\right] = \sum_{d}^{D}\left(\unmed\ln(2\pi e) + \unmed\ln|\Sigma_{\mathbf{v}}|\right) = \frac{D}{2}\ln(2\pi e) + \frac{D}{2}\ln|\Sigma_{\mathbf{v}}|
\end{equation}

\begin{equation}
    \E_q\left[\ln q_{\ys}(\mathbf{y})\right] = \sumn\left(\unmed\ln(2\pi e) + \unmed\ln|\Sigma_{\mathbf{y}}|\right)
\end{equation}

\begin{equation}
    \E_q\left[\ln q_{b}(b)\right] = \unmed\ln(2\pi e) + \unmed\ln|\Sigma_b|
\end{equation}

\begin{equation}
    \E_q\left[\ln q_{\alfas}(\bm{\delta})\right] = \sum_{d}^{D}\left(\alpha_d^{\delta} + \ln\left(\Gamma(\alpha_d^{\delta})\right) - (1 - \alpha_d^{\delta})\omega(\alpha_d^{\delta}) - \ln(\beta_d^{\delta})\right)
\end{equation}

\begin{equation}
    \E_q\left[\ln q_{\fis}(\bm{\psi})\right] = \sum_{\tilde{n}}^{\tilde{N}}\left(\alpha_{\tilde{n}}^{\psi} + \ln\left(\Gamma(\alpha_{\tilde{n}}^{\psi})\right) - (1 - \alpha_{\tilde{n}}^{\psi})\omega(\alpha_{\tilde{n}}^{\psi}) - \ln(\beta_{\tilde{n}}^{\psi})\right)
\end{equation}

\begin{equation}
    \E_q\left[\ln q_{\taus}(\mathbf{\taus})\right] = \alpha^{\taus} + \ln\left(\Gamma(\alpha^{\taus})\right) - (1 - \alpha^{\taus})\omega(\alpha^{\taus}) - \ln(\beta^{\taus})
\end{equation}

Once we have all elements of $L(q)$, we will join them and simplify (since many of them will cancel each other out) to obtain the expression of Eq. \eqref{eq:ELBO}

\appendix
\setcounter{equation}{0}
\setcounter{figure}{0}
\setcounter{table}{0}
\setcounter{section}{4}
\makeatletter
\renewcommand{\theequation}{E.\arabic{equation}}
\renewcommand{\thefigure}{E\arabic{figure}}

\section{Predictive distribution}
\label{sec_Ap3}

The posterior distribution $p(\yas|\mathbf{t},\X,\mathbf{x}_{*,:})$ is expressed as
\begin{equation}
    p(\yas|\mathbf{t},\X,\mathbf{x}_{*,:}) = \int_{\chi_{\Aa}} \int_{\chi_{\vd}} \int_{\chi_{\bb}} \int_{\chi_{\taus}} p(\yas|\mathbf{t},\X,\mathbf{x}_{*,:}, \boldsymbol{\Theta})q_{\as}\left(\textbf{a}\right)q_{\taus}(\tau)q_{\bb}(\bb)\left(\prod_d^{D}q_{\vd}(\vd)d \vd\right)d\mathbf{a}d b d\tau,
\end{equation}

Integration over noise $\taus$ is intractable; however, as \cite{bishop2013variational} shows, the gamma distribution of $\taus$ becomes concentrated around its mean as the number of samples increases. It is due to the asymptotic behavior of the variance of a Gamma distribution as $\langle \taus^2 \rangle - \langle \taus \rangle^2 = \frac{a}{b^2} \sim O(\frac{1}{N})$ (the variance tends to 0 as $N$ increases). So, we could approximate (E.1) as:
\begin{equation}
    p(\yas|\mathbf{t},\X,\mathbf{x}_{*,:}) \simeq \int_{\chi_{\Aa}} \int_{\chi_{\vd}} \int_{\chi_{\bb}} p(\yas|\mathbf{t},\X,\mathbf{x}_{*,:}, \boldsymbol{\Theta}_{-\taus},\tauE)q_{\as}\left(\textbf{a}\right)q_{\bb}(\bb)\left(\prod_d^{D}q_{\vd}(\vd)d \vd\right)d\mathbf{a} d b.
\end{equation}

Thus, we can express the posterior predictive distribution as:
\begin{equation}
\label{eq:CCC}
\begin{split}
    p(\yas|\mathbf{t},\X,\mathbf{x}_{*,:}) \simeq & \int_{\chi_{\Aa}} \int_{\chi_{\vd}} \int_{\chi_{\bb}} \N(\mathbf{x}_{*,:}\vlam\Xst\as + b, \tauE^{-1})\N(\aE, \acov)d\mathbf{a}\left(\prod_d^{D}\N_F(\langle \vd \rangle, \sigma_{d})d \vd\right) \\ 
    & \N(\langle b \rangle,\Sigma_{b})d b,
\end{split}
\end{equation}
which is the convolution of some gaussian probability density functions that we can solve by operating the multiple integral step-by-step. Hence, we can rewrite the Eq. (\ref{eq:CCC}) as
\begin{equation}
\begin{split}
    p(\yas|\mathbf{t},\X,\mathbf{x}_{*,:}) \simeq & \int_{\chi_{\vd}} \bigg[\int_{\chi_{\bb}} \left[\int_{\chi_{\Aa}} \left[\N(\mathbf{x}_{*,:}\vlam\Xst\as + b, \tauE^{-1})\N(\aE, \acov)d\A\right]\N(\langle b \rangle,\Sigma_{b})d b\right]\\ & \prod_d^{D}\N_F(\langle \vd \rangle, \sigma_{d})d \vd\bigg].
\end{split}
\end{equation}

The integral defined over $\A$, can be solved as a convolution of gaussian densities, leaving the expression as
\begin{equation}
\begin{split}
    p(\yas|\mathbf{t},\X,\mathbf{x}_{*,:}) \simeq & \int_{\chi_{\vd}} \bigg[ \int_{\chi_{\bb}} \bigg[\N(\mathbf{x}_{*,:}\vlam\Xst\aE + b, \tauE^{-1} + \mathbf{x}_{*,:}\vlam\Xst\Sigma_{\A}\Xs\vlam\mathbf{x}_{*,:}^\top) \\ 
    & \N(\langle b \rangle,\Sigma_{b})d b\bigg] 
     \prod_d^{D}\N_F(\langle \vd \rangle, \sigma_{d})d \vd\bigg].
\end{split}
\end{equation}

Following the same procedure, we can simplify the term $b$ as
\begin{equation}
\begin{split}
    p(\yas|\mathbf{t},\X,\mathbf{x}_{*,:}) \simeq & \int_{\chi_{\vd}} \bigg[ \N(\mathbf{x}_{*,:}\vlam\Xst\aE + \langle b\rangle, \tauE^{-1} + \mathbf{x}_{*,:}\vlam\Xst\Sigma_{\A}\Xs\vlam\mathbf{x}_{*,:}^\top + \Sigma_{b}) \\ & \prod_d^{D}\N_F(\langle \vd \rangle, \sigma_{d})d \vd\bigg].
    \label{eq:pred_fin}
\end{split}
\end{equation}

Finally, we must take into account the $\N_F$ distribution of each $\vd$. The normal folded distribution is defined only over positive values, which will affect the range of integration; also, it is not a symmetric distribution around its mean. Furthermore, $p(\yas|\mathbf{t},\X,\mathbf{x}_{*,:}, \mathbf{v})$ presents both at mean and variance the variable $\mathbf{v}$, making intractable the marginalization over $\mathbf{v}$.

Hence, we know that the integral of Eq. \eqref{eq:pred_fin} can be treated as a convolution of two probabilistic distributions of known shape. Thus, we can extract some properties from the final distribution without solving the integral. First, the final distribution must have only one maximum in its domain; this is due to the fact that the convolution of two unimodal functions in the real domain always generates a unimodal function \cite{ibragimov1956composition}. Second, the final function will not be symmetric around its mean; as the convolution of one symmetric distribution and one asymmetric distribution will always lead to an asymmetric distribution.

In view of the above, we can directly estimate the mean and variance of $p(\yas|\y,\X,\mathbf{x}_{*,:})$. Thus, as the mean-field procedure assumes independence between all model variables, we can calculate the expectation of $\y$ as:
\begin{equation}
    \E[\yas] = \E[\mathbf{x_{*,:}}\vlam\Xst\as + b + \epsilon] = \mathbf{x_{*,:}}\vlamE\Xst\aE + \langle b\rangle.
\end{equation}

Also, considering the mean-field independence, it is easy to determine the variance of $\yas$ as:
\begin{equation}
    Var[\yas] = Var[\mathbf{x_{*,:}}\vlam\Xst\as + b + \epsilon] = Var[\mathbf{x_{*,:}}\vlam\Xst\as] + \Sigma_{b} + \tauE,
    \label{eq:var_y}
\end{equation}
where we can express the first term as:
\begin{equation}
    Var[\mathbf{x_{*,:}}\vlam\Xst\as] = Var\left[\sum_d^{D}(x_{*,d} \vd \xsdt\as)\right] = \sum_d^{D}\left(x_{*,d}^2 \xsdt Var[\vd \as]\xsd\right).
\end{equation}

Moreover, we can develop the variance of $\vd \as$ as:
\begin{equation}
\begin{split}
    Var[\vd \as] &= \E[\vd^2]\E[\A^T \A] - \E[\vd]^2\E[\A]^2 = (Var[\vd] + \E[\vd]^2)(Var[\A] + \E[\A]^2) - \E[\vd]^2\E[\A]^2\\
    & = Var[\A]Var[\vd] + \E[\A]^2Var[\vd] + \E[\vd]^2Var[\A] = \Sigma_{\A}\sigma_{\vd} + \aE^T\aE\sigma_{\vd} + \langle \vd \rangle^2\Sigma_{\A}.
    \label{eq:des_var}
\end{split}
\end{equation}

Thus, by joining Eq. \eqref{eq:des_var} and Eq. \eqref{eq:var_y}, we can express the resulting variance of $\yas$ as:
\begin{equation}
\begin{split}
    \zeta^2_{\yas} & = \mathbf{x}_{*,:}\vlamE\Xst\Sigma_{\A}\Xs\vlamE\mathbf{x}_{*,:}^\top + \tauE + \Sigma_{b} + \sum_d^{D} \left(x_{*,d}^2 \xsdt(\aE^\top\aE\sigma_{\vd} + \Sigma_{\A}\sigma_{\vd})\xsd \right)\\
    & = \mathbf{x}_{*,:}\vlamE\Xst\Sigma_{\A}\Xs\vlamE\mathbf{x}_{*,:}^\top + \tauE + \Sigma_{b} + \sum_d^{D} \left(x_{*,d}^2 \xsdt\sigma_{\vd}\langle \A^\top\A\rangle\xsd \right).
\end{split}
\end{equation}

\newpage

\bibliographystyle{plain}
\bibliography{refs.bib}

\end{document}